\documentclass[journal=jacsat,manuscript=article]{achemso}

\usepackage[version=3]{mhchem} 
\usepackage{float}
\usepackage{hyperref}
\usepackage{booktabs}
\usepackage{tabularx}
\usepackage{array}
\usepackage[numbers]{natbib} 

\newcommand{\mycaption}[2]{\caption{\textbf{#1}. #2}}

\author{Jose Manuel Barraza-Chavez}
\affiliation{Department of Chemical Engineering and Applied Chemistry, University of Toronto, Toronto}
\alsoaffiliation{Vector Institute, Toronto}

\author{Rana A. Barghout}
\affiliation{Department of Chemical Engineering and Applied Chemistry, University of Toronto, Toronto}
\alsoaffiliation{Vector Institute, Toronto}

\author{Ricardo Almada-Monter}

\author{Adrian Jinich}
\affiliation{University of California, Skaggs School of Pharmacy and Pharmaceutical Studies, San Diego}

\author{Radhakrishnan Mahadevan}
\affiliation{Department of Chemical Engineering and Applied Chemistry, University of Toronto, Toronto}

\author{Benjamin Sanchez-Lengeling}
\email{ben.sanchez@utoronto.ca;jose.barrazachavez@mail.utoronto.ca} 
\affiliation{Department of Chemical Engineering and Applied Chemistry, University of Toronto, Toronto}
\alsoaffiliation{Vector Institute, Toronto}

\title[An \textsf{achemso} demo]
  {Graph Data Modeling:\\ Molecules, Proteins, \& Chemical Processes}

\abbreviations{IR,NMR,UV}

\usepackage{placeins}
\begin{document}

\subparagraph{}\textit{\small \centering This document is the unedited Author’s version of a Submitted Work that was subsequently accepted for publication in \href{https://pubs.acs.org/series/infocus}{ACS In Focus}, copyright ©2025 after peer review. The \href{https://pubs.acs.org/doi/book/10.1021/acsinfocus.7e9017}{final published work} will be available September 2025 as part of the ACS in Focus  two-to-four-hour primer reads.}

\begin{figure}[!htbp]
  \centering
  \includegraphics[width=1.0\columnwidth]{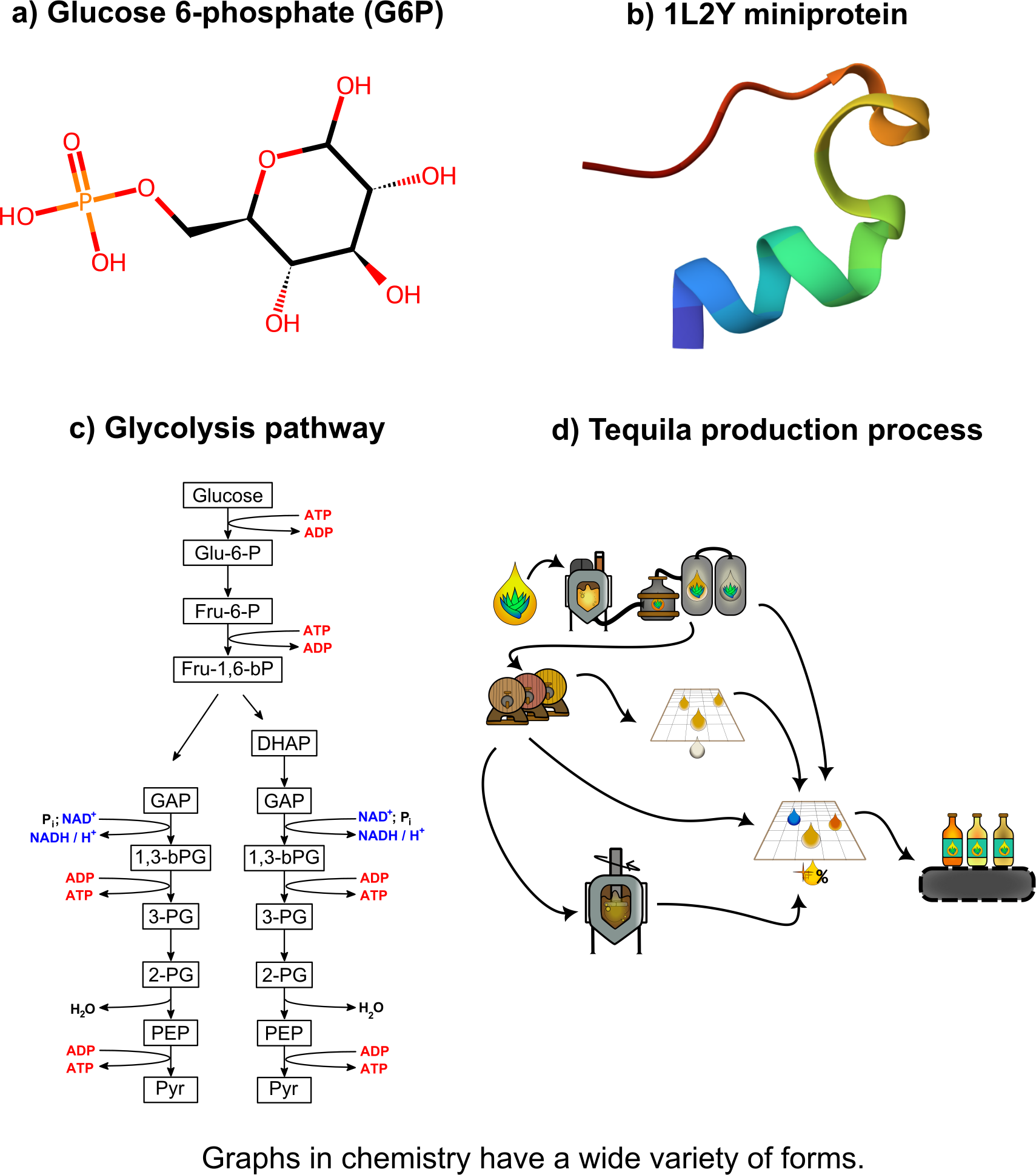}
  \mycaption{Examples of graphs you might find in chemical fields}{a) Molecular graph of glucose 6-phosphate, b) Protein graph of 1L2Y miniprotein \cite{1}, c) Reaction graph of the glycolysis metabolic pathway, d) Industrial chemical process graph of the tequila production process.}
  \label{Figure1}
\end{figure}

\paragraph{Graphs: the language of interactions} Graphs are all around us, especially if you are in chemical sciences. Figure \ref{Figure1} has four distinct graphs, and you are probably already familiar with one or more of them. Let’s take glucose 6-phosphate (G6P), a key metabolite in glycolysis, as an example. While the molecule represents a material object in 3D space, we often choose to represent it as a molecular graph, with atoms as nodes and covalent bonds as edges. Molecular graphs are a powerful abstraction to represent chemical phenomena. Trained chemists are often able to infer function or the relationship to other molecules (reactions) by inspecting a graph. With algorithms and data we can start to do the same. This primer is an introduction to graphs as a mathematical object in the context of chemical data and how we can use learning algorithms, more specifically, graph neural networks (GNNs) to build systems that operate on graphs: predicting their function, generating or manipulating graphs.
Just like graphs, molecules are all around us, impacting everything from the materials we use to the medicine we take. As chemistry (and chemistry-adjacent fields) increasingly embraces data-driven approaches, the ability to work with graphs and model them is becoming essential. Researchers are already using these tools to discover novel antibiotics and odors, design materials with tailored properties, uncover unexplored areas of metabolism and optimize chemical processes. By mastering the concepts presented here, you'll be prepared to contribute to the next generation of chemical breakthroughs.
Our first section covers the mathematics and design of graphs, while in the second section, we formulate many prediction problems as graph-based tasks. We provide more examples of graphs in section three. Finally, in the fourth section, we look at the modeling space of learning algorithms for graphs with an emphasis on graph neural networks.
\FloatBarrier

\section{1. Graphs and graph data}
\begin{figure}[!htb] \centering
  \centering
\includegraphics[width=1.0\columnwidth]{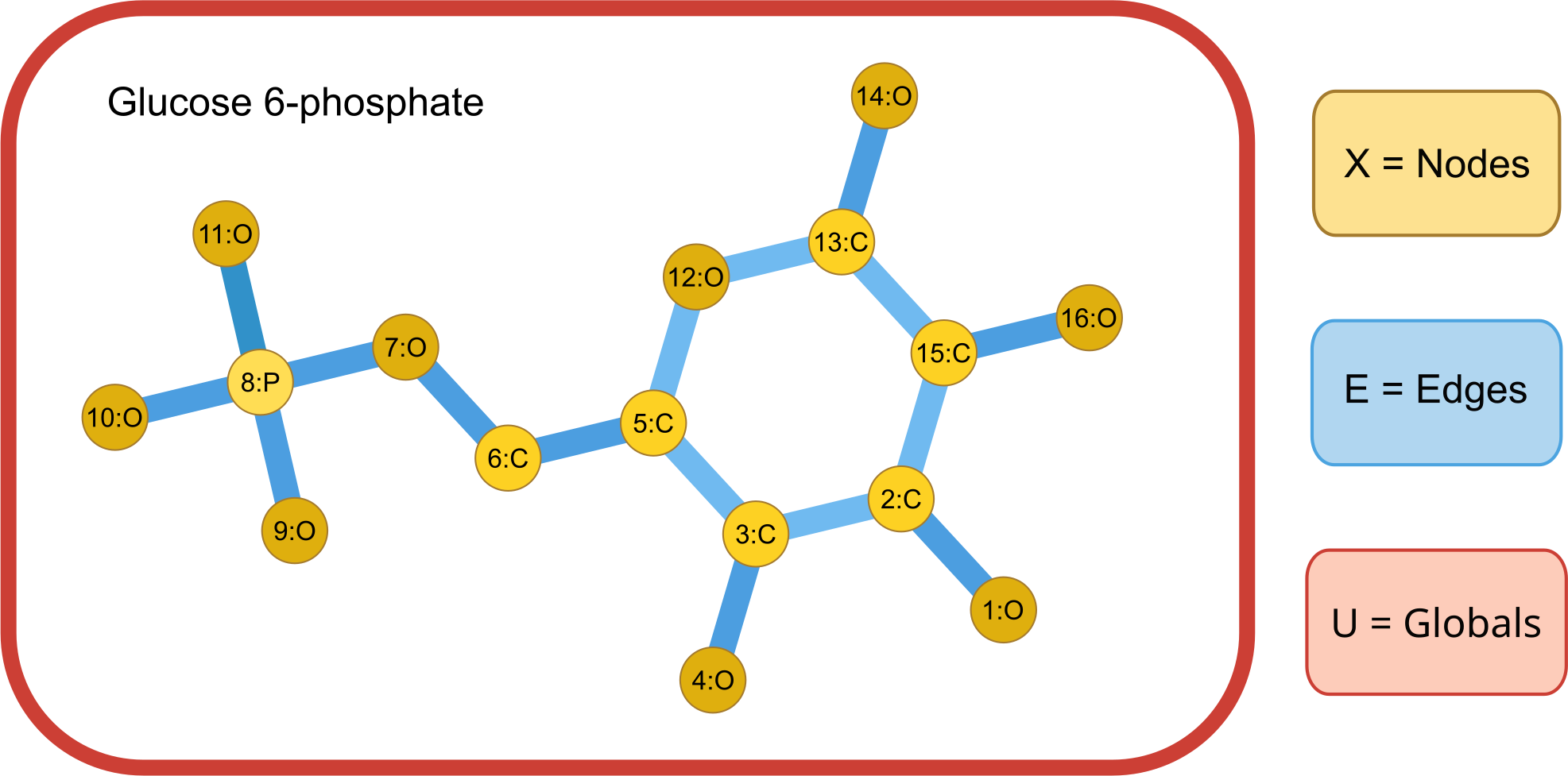}
  \mycaption{The components of a graph (nodes, edges, globals)}{Using glucose-6P as an example graph, highlighting each component and establishing the graph coloring scheme.
  }\label{Figure2}
\end{figure}

Graphs are a mathematical structure that represents relationships between objects. They are especially good at representing chemical and biological science phenomena as these fields tend to study individual components (like cells, molecules, proteins) and their interactions (bonds, molecular interactions, cell signaling). Graphs come in a variety of shapes and sizes; however, their main components are presented in Figure \ref{Figure2}, and are as follows:

\textbf{Nodes:} Represent individual objects. Also called vertices. 

\textbf{Edges:} Represent the relationship that exists between two objects and represent the flow of information between nodes. Also called links or connections. Edges can have directionality: they can be undirected or directed.

\textbf{Global:} Refers to properties inherent to the entire graph, also called context or master node. It can be interpreted as a “special” node that is connected to all nodes/edges \cite{2}.

\FloatBarrier

\subsection{Representing graphs numerically}
\begin{figure}[!htb] \centering
\includegraphics[width=0.8\columnwidth]{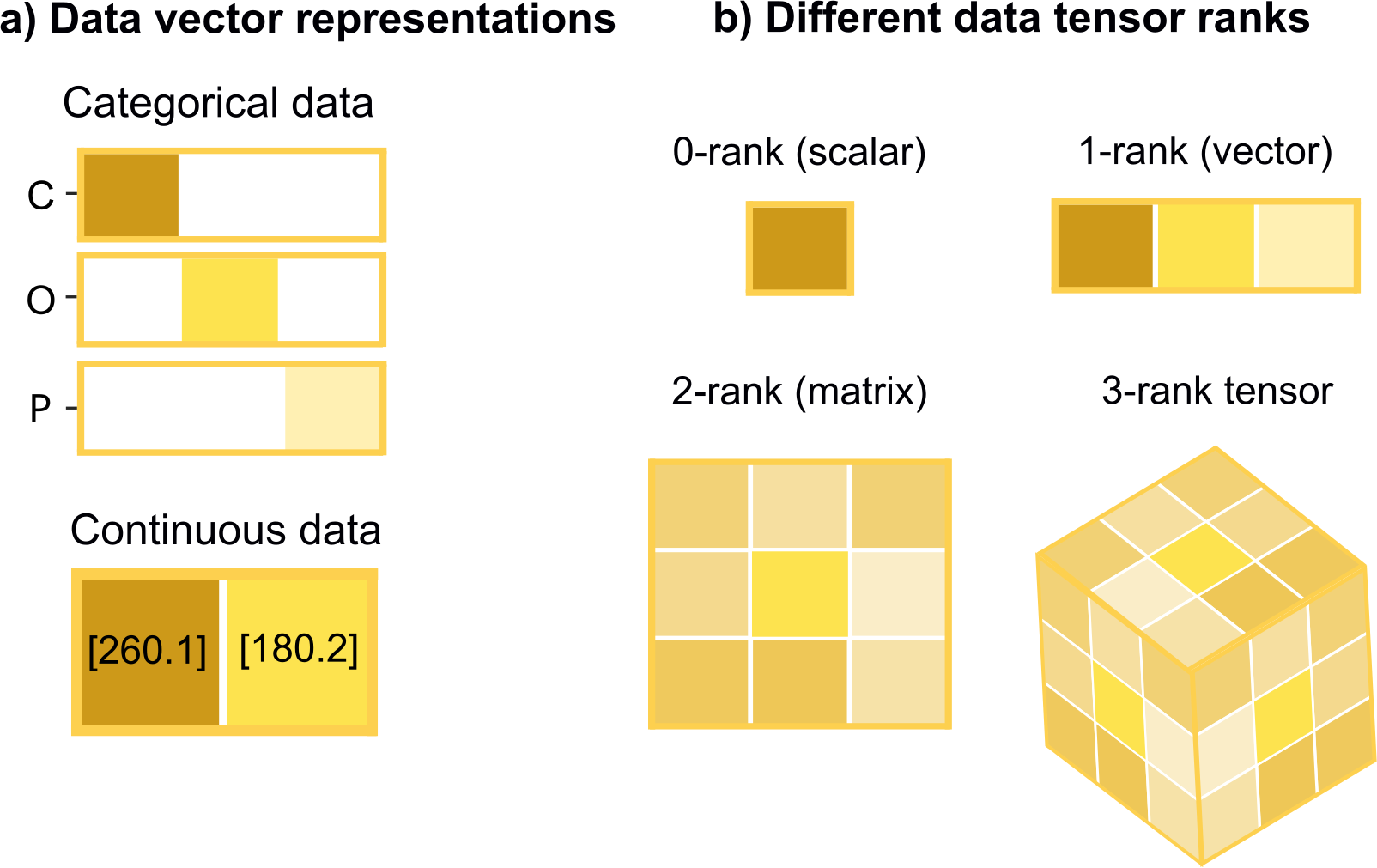}
  \mycaption{Representing data numerically}{a) Encoding categorical and continuous data into a vector b) Tensors of different ranks.
  }\label{Figure3}
\end{figure}

To manipulate data in a computer, we need to represent it numerically. Tensors, which are n-dimensional arrays, offer a flexible means of organizing data. Scalars are rank-0 tensors, vectors are rank-1, matrices are rank-2, and so on. When the data is continuous, this is straightforward; we can represent the values directly as numbers in a tensor or encode them using a graded color scale, where different shades of the same color correspond to different values (Figure \ref{Figure3}). When the data is discrete, we need to encode the data so that it can be represented numerically. If the data is categorical, we can one-hot encode it: we assign integer values to each class and then represent each instance as a binary vector where all elements are zero except for the index corresponding to its class, which is set to one (Figure \ref{Figure3}b).
For example, for a molecule like glucose 6-phosphate, we can extract information like the type of atom (“oxygen”, “carbon”, “phosphorus”), as well as the molar mass of each atom. We can one-hot encode these categories (atom type) into a 3-dimensional vector, for example carbon will correspond to [1, 0, 0]. When repeated to all nodes, we obtain a rank-2 tensor with the number of rows equal to the number of nodes and three columns, each corresponding to an atom type.
Similarly, if we consider a covalent bond edge and want to represent it numerically, we can do so by using one-hot encoding for bond types like "single" or "double", and a binary encoding to indicate whether the bond is cyclic as true or false with 1 or 0. These values can then be concatenated into a 3-dimensional vector. If we do this for all edges, we get a rank-2 tensor as well.

\begin{figure}[!htb] \centering
  \includegraphics[width=0.8\columnwidth]{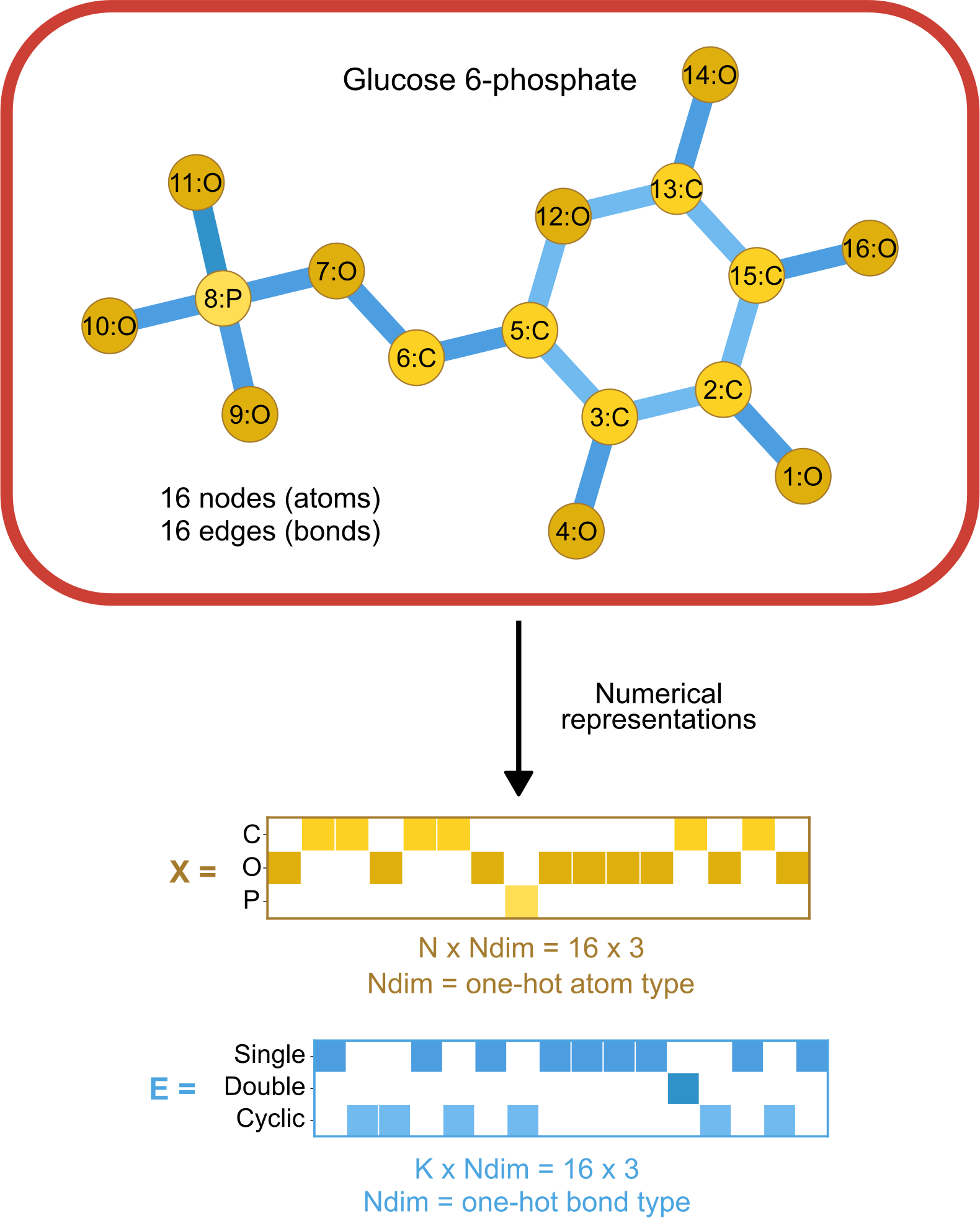}
  \mycaption{G6P represented as a graph tensor}{Showcasing node (X) and edge (E) tensors.
  }\label{Figure4}
\end{figure}

Considering a graph has many components: nodes, edges and globals, we can represent a graph numerically as a collection of tensors, one for each component. Figure \ref{Figure4} showcases one example of such a collection for G6P.

\FloatBarrier

\subsection{Representing the Connectivity of a Graph}

\begin{figure}[!htb] \centering
\includegraphics[width=0.5\columnwidth]{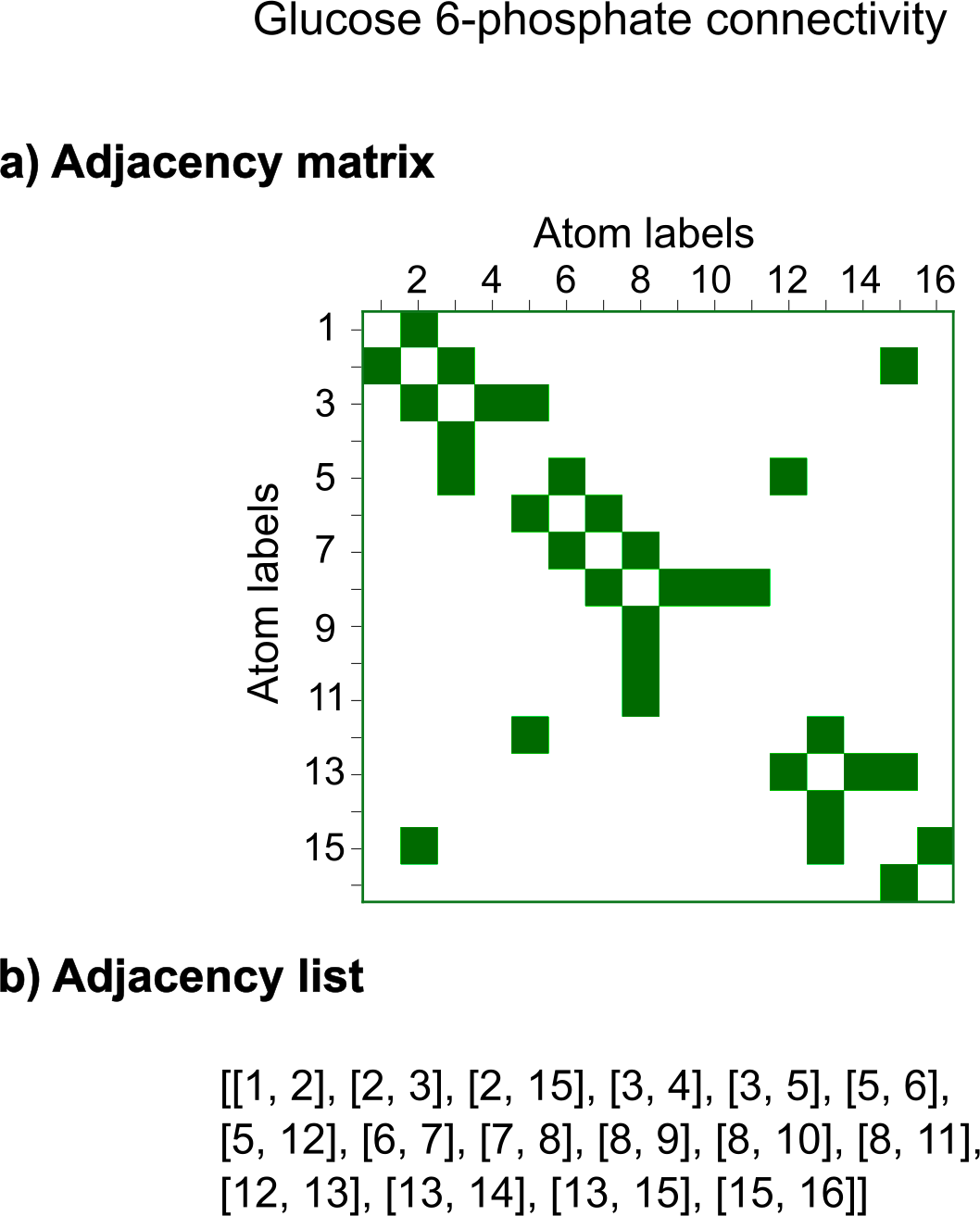}
  \mycaption{Representing connectivity of a graph as matrix and as a list}{a) Adjacency matrix for G6P b) Adjacency list for G6P.
  }\label{Figure5}
\end{figure}

So far, we have seen how to represent the information that is contained in each element of the graph, but not how they are connected. To represent the connectivity information of a graph, we can use adjacency matrices or adjacency lists (Figure \ref{Figure5}). An adjacency matrix is a grid where each entry represents the presence or absence of a connection between two nodes. For example, if there is an edge between node b and node a, the corresponding entry in the matrix will be 1; otherwise, it will be 0.
Adjacency matrices, while straightforward, tend to be very sparse, especially in large graphs where there are significantly more unconnected nodes than connected ones. This sparsity leads to inefficiencies: a graph with n nodes requires an adjacency matrix with n x n entries, most of which are zeros. For example, representing a protein like hexokinase (the first enzyme in G6P catalysis) as a molecular graph, with approximately 15,000 atoms, would require an adjacency matrix of size 15,000 x 15,000. This large size makes adjacency matrices computationally expensive for large-scale graphs. In contrast, adjacency lists avoid this inefficiency by recording only the existing edges, making them better suited for sparse graphs.
In an adjacency list (also called edge list), we assign indices to each node and represent edges as pairs of node indices they connect. For example, in a graph with four nodes A—B—C—D, indexed as [1, 2, 3, 4], the edge list would be: (1,2), (2,3), (3,4).
If a graph is undirected, the adjacency matrix will be symmetric, where each edge goes both ways A to B and B to A. In undirected graphs, the adjacency list can represent each edge only once, since the connection is bidirectional. In a directed graph, we will not observe this symmetry. Some graphs are considered weighted; they have weights (scalars) associated with each connection. These are just special cases of edge information and can be included in edge tensor information or even in weighted adjacency matrices (Figure \ref{Figure9}) or lists. Later in this section, we will see an example of a weighted graph and how this information is portrayed in a tensor.

\FloatBarrier

\subsection{Types of Graphs}

\begin{figure}[!htb] \centering
  \includegraphics[width=0.8\columnwidth]{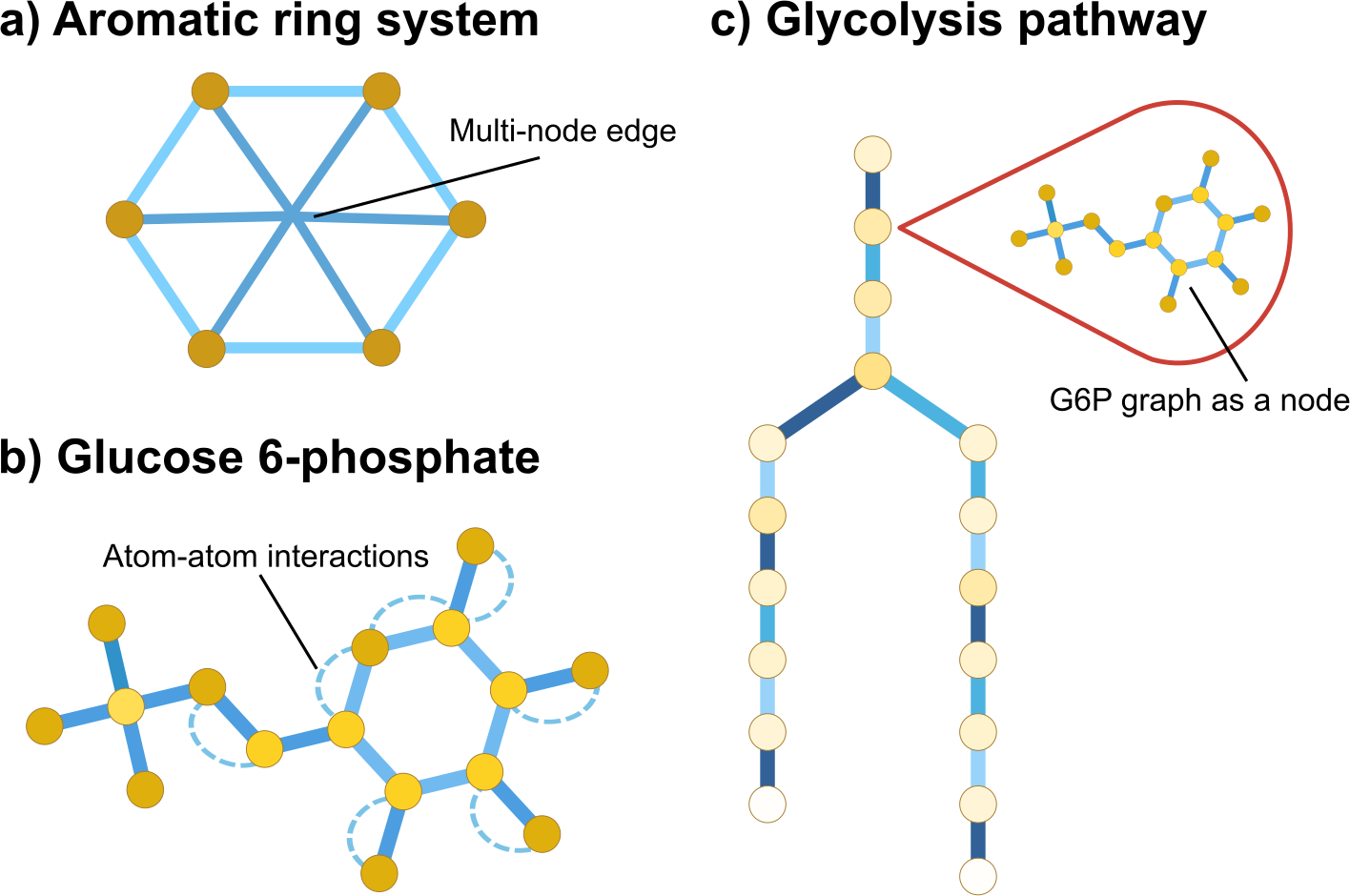}
  \mycaption{Different types of nodes and edges in heterogeneous graphs}{a) Ring system as a multi-node edge b) Multiple edges that connect the same nodes (atom-atom interactions) c) Glycolysis graph as a hierarchical graph.
  }\label{Figure6}
\end{figure}

Chemistry and biology often involve many parts interacting in complex ways, at different scales. Graphs are flexible and they allow for different types of interactions: we can have an edge that connects multiple nodes or a node that has multiple edges. For example, when considering the molecular graph (G6P), we can think of a cyclic system as an interaction (edge) that involves multiple atoms (nodes) as seen in Figure \ref{Figure6}a, or if we specify other types of bonds (hydrogen, ionic, charge-shift, dispersion), a node can even have multiple edges as seen in Figure \ref{Figure6}b. As such, we can classify two general types of graphs: homogeneous and heterogeneous. Homogeneous graphs have a single type of node and edge, while heterogeneous graphs have many types of edges and nodes \cite{2}. 
This definition is very broad, allowing even hierarchical graphs, where a node is also a graph itself. Let’s consider the glycolysis pathway graph (Figure \ref{Figure6}c), which is a homogeneous graph if we consider its nodes as molecules and edges as enzymatic reactions. As we saw with G6P, we can also represent molecules as homogeneous graphs (nodes, edges). Using the molecular graph as a node in the pathway graph, we construct a hierarchical (and heterogeneous) graph with two types of nodes (atoms, molecules) and two types of edges (covalent bonds, enzymatic reactions).

\FloatBarrier

\subsection{Building graphs, step by step}
Graphs are flexible abstractions of relational data. The information we express with them depends on how we want to use them. To build good graph representations of chemical systems, it's important to think about all the different parts of a graph, how they relate to each other, and the choices we make to express them. Let's make this more concrete by revisiting the four graph examples in the first figure.

\subsection{Building a molecular graph: glucose 6-phosphate}
\begin{figure}[!htb] \centering
  \includegraphics[width=1.0\columnwidth]{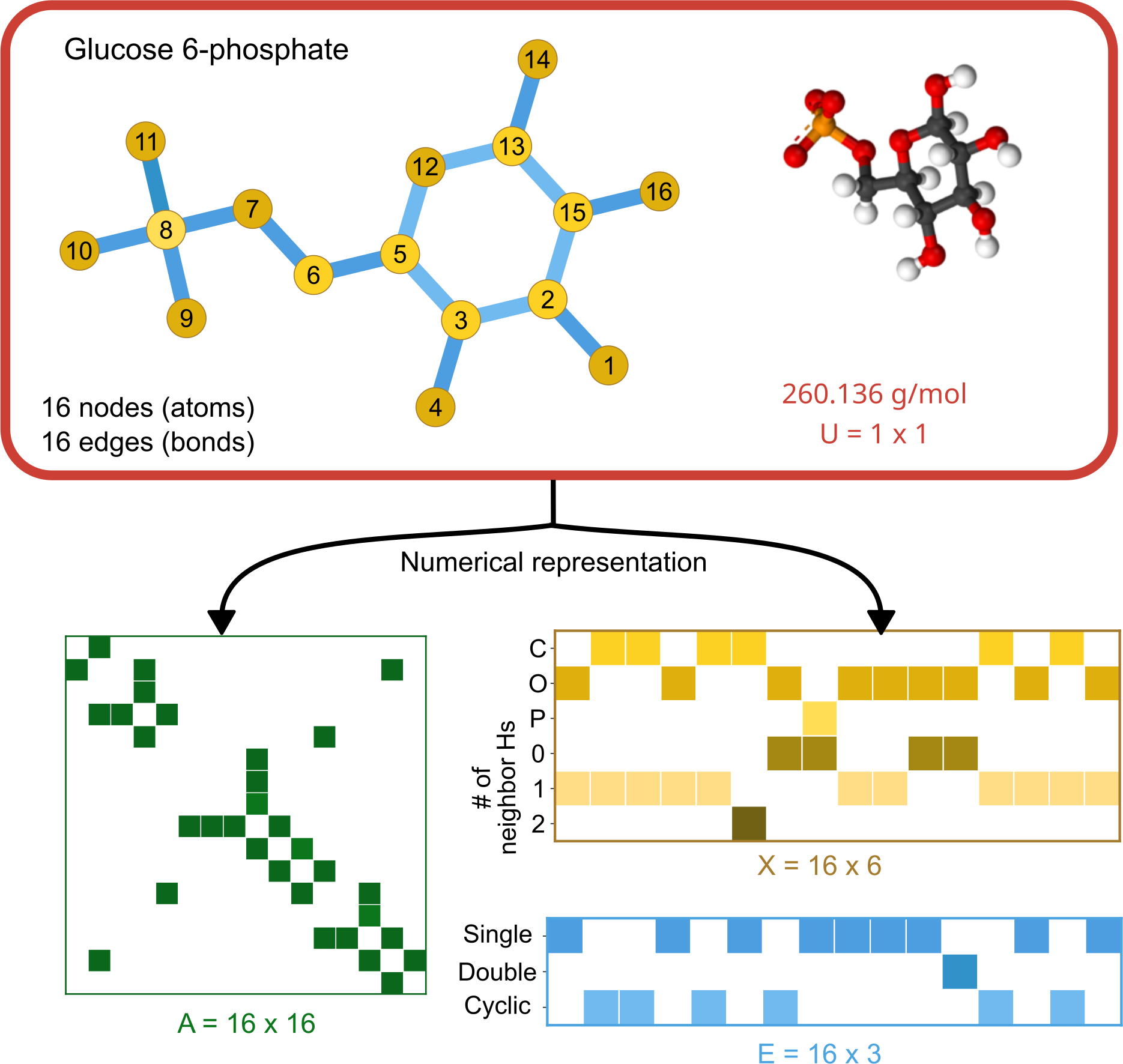}
  \mycaption{Graph tensor for G6P}{Showcase of adjacency matrix (A), node tensor (X) with dimensions 16 nodes x 6 where atom type and number of hydrogens is one-hot-encoded as features, edge tensor (E) with dimension 16 edges x 3 one-hot encoded bond type as features, and global tensor (U).
  }\label{Figure7}
\end{figure}

As we’ve mentioned before, the most common chemical graph is a molecular graph, where non-hydrogen atoms are nodes and covalent bonds are edges. For this G6P abstraction, we have 16 nodes and 16 edges. How do we represent this graph as a (structured) tensor? In a minimal setting, we specify the identity of the atoms (C, O, P) and one-hot encode them into a 3-dimensional vector (as shown in Figure \ref{Figure7}). We can also one-hot encode the number of hydrogens connected to the atom (0-2) as a 3-dimensional vector. Therefore, the collection of all atom vectors in our 2D node tensor is of shape 16 x 6. In the same way, we can represent the covalent identity of each bond (single, double) resulting in a 2D edge tensor of 16 x 2, if we add information indicating whether the bond is in a cyclic system (true or false), we obtain an edge tensor of 16 x 3. We could also specify a global property of the molecule, such as molecular weight as a scalar in a 1D tensor (1 x 1). These three tensors represent (some) information that describes G6P, together with the adjacency matrix (16 x 16), we have a collection of 4 tensors (Figure \ref{Figure7}).

\FloatBarrier

\begin{figure}[!htb] \centering
  \includegraphics[width=0.8\columnwidth]{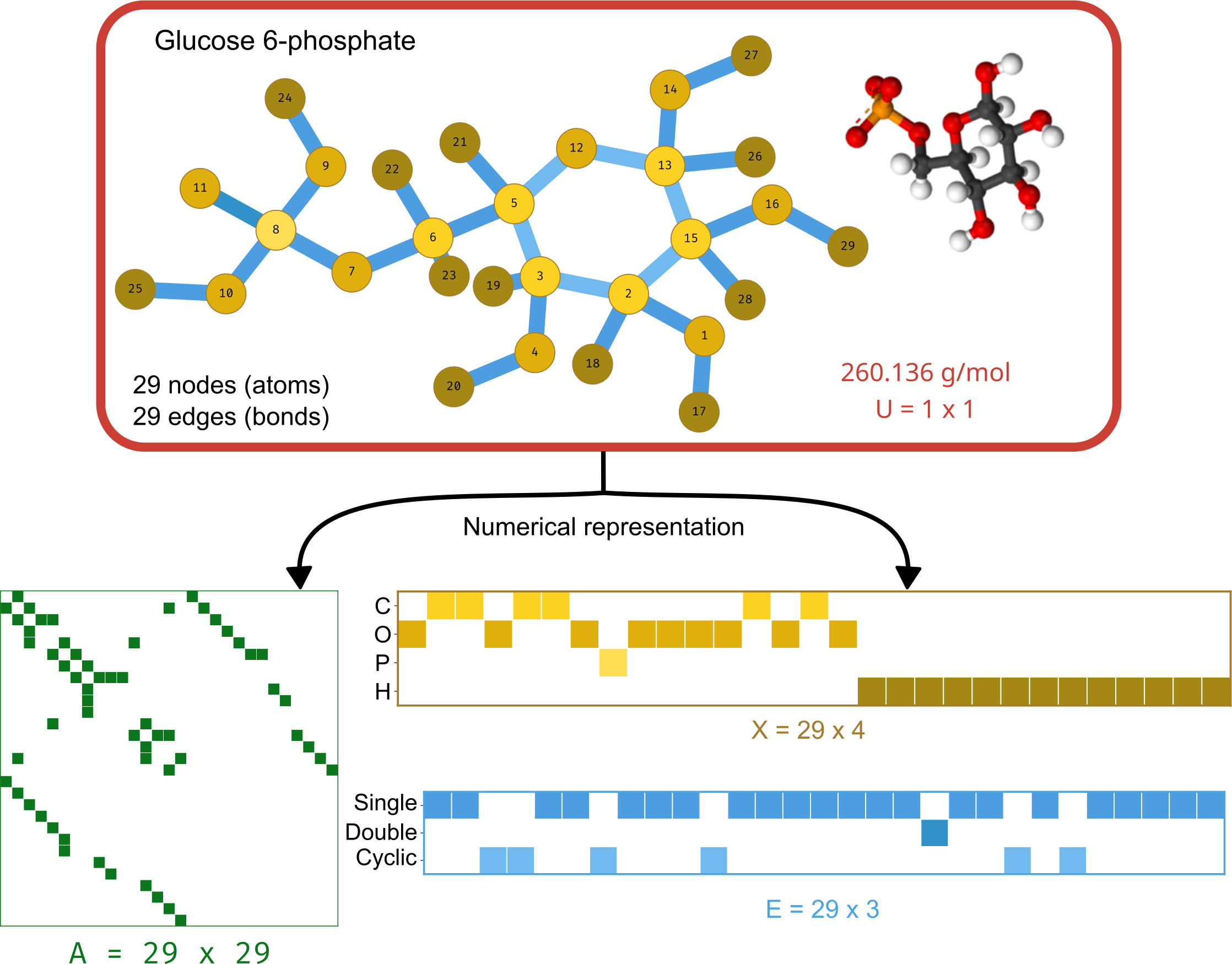}
  \mycaption{Graph tensor for G6P with explicit hydrogens}{Showcase of adjacency matrix (A), node tensor (X) with dimensions 29 nodes x 4 one-hot encoded atom type as features, edge tensor (E) with dimensions 29 edges x 3 one-hot encoded bond type as features, and global tensor (U).
  }\label{Figure8}
\end{figure}

This graph is an abstraction; it’s a useful way to describe a molecule but it's not the only one. For example, we could explicitly express the hydrogens as nodes in G6P. Our graph would contain 29 nodes and 29 edges, in this new representation (Figure \ref{Figure8}). We can one-hot encode each atom type (C, H, O, P) to a 4-dimensional vector, where one-hot encoding the number of hydrogens as a categorical feature is no longer necessary, since it is included in the number of nodes, leaving us with an atom tensor of 29 x 4. Our edge tensor is now a 29 x 3 array. The adjacency matrix is 29 x 29.

\FloatBarrier

The dimensionality of the node, edge, global vectors will depend on how much information we want to express. For example, we could augment information for each component by adding implicit charge to the nodes, stereochemistry information to the edges, and aqueous solubility (logS) to the global vector. Typically, more interaction rich (heterogeneous graphs) or information (node/edge/global) rich graphs will result in higher performing models at a cost of modelling complexity (further explored in section 3).

\subsection{Building a protein graph}

\begin{figure}[!htb] \centering
  \includegraphics[width=1.0\columnwidth]{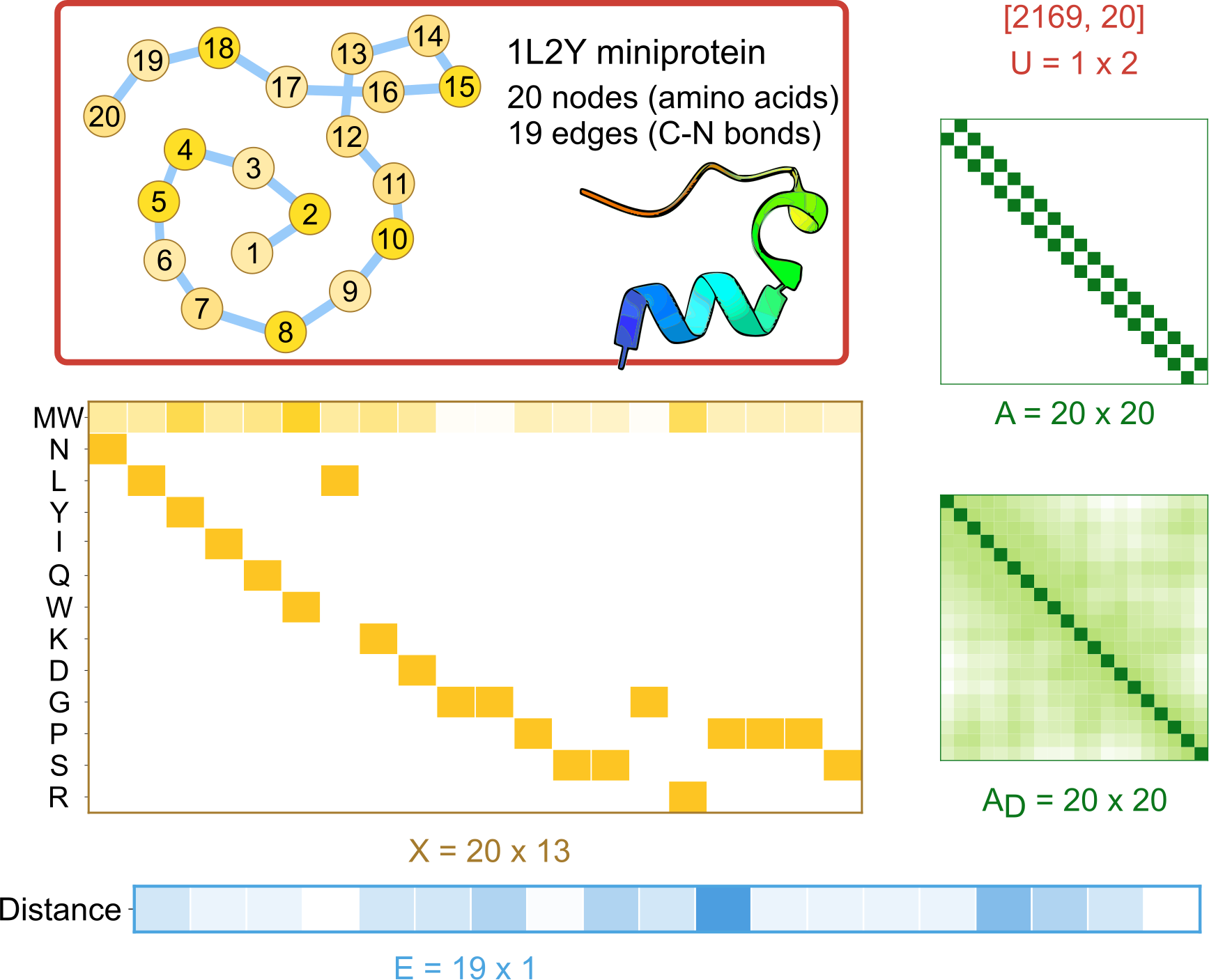}
  \mycaption{Graph tensor for 1L2Y}{Showcase of adjacency matrix (A), distance matrix (AD), node tensor (X) with dimensions 20 nodes x 13 one-hot encoded amino acid type as features, edge tensor (E) with dimensions 19 edges x 1 distance feature, and global tensor (U).
  }\label{Figure9}
\end{figure}

On a bigger scale, we can model large molecules like proteins as graphs. In this example we will work with the 1L2Y miniprotein, a small, 20-residue, highly stable, and well-characterized protein designed to study protein folding. In this graph representation, we can consider the amino acid residues as nodes and the covalent bonds that unite them as edges. As globals, we can extract the total molar mass of the protein, as well as the number of amino acid residues. For the nodes, we can utilize the molar mass of each amino acid and the one-hot encoded amino acid type as features; because this 1L2Y has only 12 unique amino acids, the resulting node tensor has dimension 20 x 13 (Figure \ref{Figure9}). Similarly for the edges, we can extract information about the distances of each covalent bond, yielding a 19 x 1 tensor. In addition, we can obtain an adjacency distance matrix that acts as a heat map representing every value of distance between every node similar to a protein contact map, this is an example of a weighted adjacency matrix. This distance matrix is a tensor with dimension 20 x 20. Trivially, we can also add the basic adjacency matrix for each covalent C-N bond between the amino acid residues, giving us another 20 x 20 tensor. In total we will have 5 tensors for this graph, one for nodes, one for edges, one for the global and 2 different adjacency matrices.

\FloatBarrier

\begin{figure}[!htb] \centering
  \includegraphics[width=1.0\columnwidth]{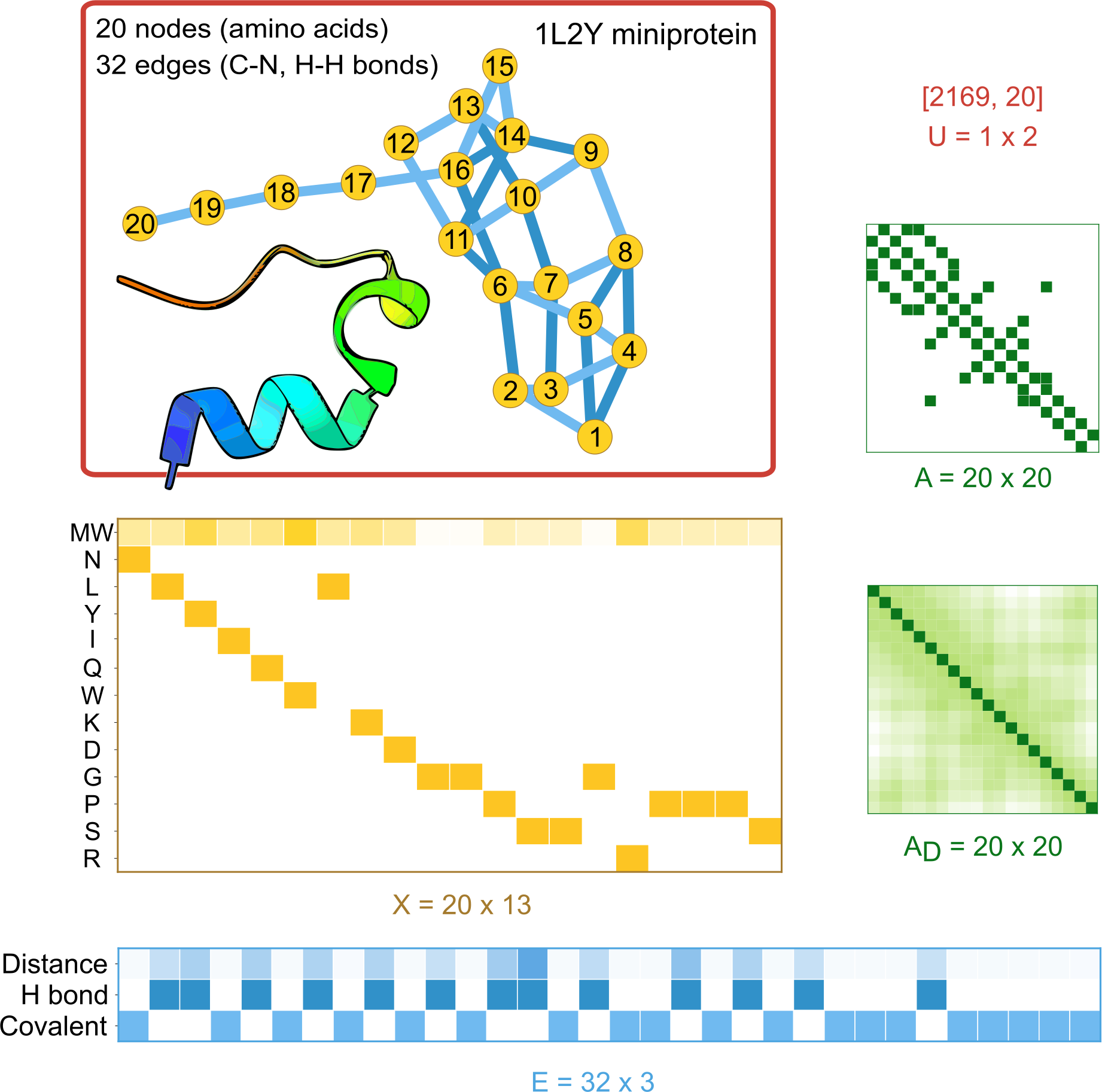}
  \mycaption{Graph tensor for 1L2Y with hydrogen bonds}{Showcase of adjacency matrix (A), distance matrix (AD), node tensor (X) with dimensions 20 nodes x 13 one-hot encoded amino acid type as features, edge tensor (E) with dimensions 32 edges x 3 distance feature, and global tensor (U). 
  }\label{Figure10}
\end{figure}

Repeating the same exercise as before, we can construct a new graph-based abstraction for 1L2Y, where the edges are now hydrogen bonds. 1L2Y has 13 hydrogen bonds that curl the amino acid chain and shape its secondary structure, as depicted in Figure \ref{Figure10}. To integrate this information into our graph tensor, we can modify the distance and adjacency matrices to account for these hydrogen bonds, as well as the edge feature tensor. Because now we have 13 more edges, the adjacency and distance matrices will contain more entries while preserving the same dimensions, whereas the only tensor that will change dimension will be the edge feature tensor, as we can one-hot encode the bonds as covalent or hydrogen type, yielding a new edge tensor of dimensions 32 x 3.

\FloatBarrier

Having established a graph representation for individual proteins, we now shift our focus to representing entire biochemical pathways. Enzymatic proteins, which catalyze reactions within these pathways, provide a natural bridge to exploring how pathways themselves can be modeled as graphs. This representation introduces the kinds of considerations we can do while working with biochemical systems to construct and analyze graphs.

\subsection{Building a reaction graph: glycolysis pathway}

\begin{figure}[!htb] \centering
  \includegraphics[width=0.75\columnwidth]{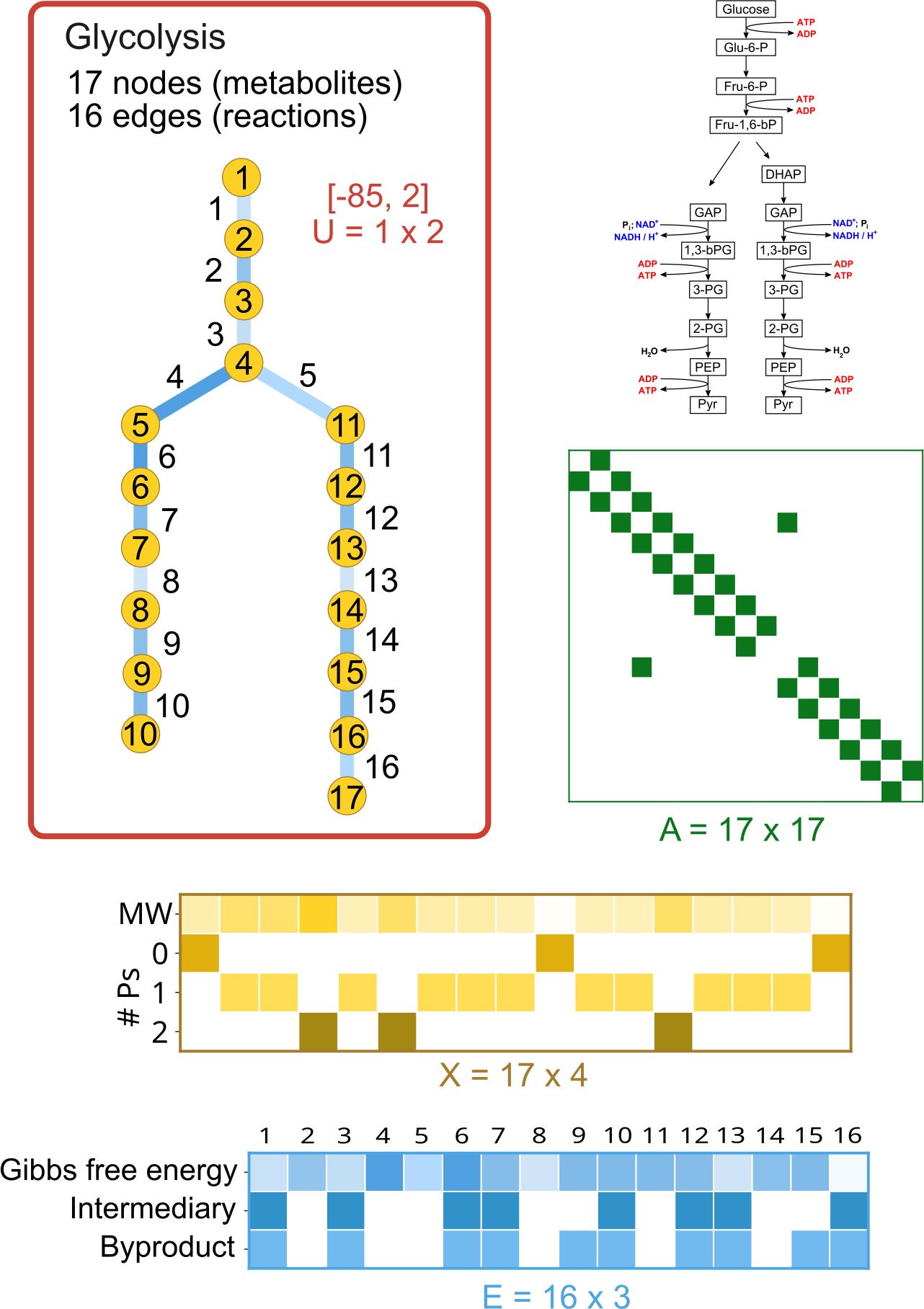}
  \mycaption{Graph tensor for glycolysis}{Showcase of adjacency matrix (A), node tensor (X) with dimensions 17 nodes x 4 features, edge tensor (E) with dimensions 16 edges x 3 features, and global tensor (U).
  }\label{Figure11}
\end{figure}

We can visualize biochemical pathways like glycolysis as reaction graphs. In this type of graph, each node represents a molecule, and each edge represents a reaction that transforms one molecule into another. You might have seen glycolysis represented as a single chain reaction of steps. However, for this first representation, we include the aldolase-catalyzed split of fructose 1,6-bisphosphate into two molecules, resulting in 17 nodes and 16 edges making up the graph. In the same way as G6P, let's represent glycolysis as a collection of tensors; let’s say we want to predict if this is an exergonic or endergonic process. For nodes, we can extract the amount of phosphate groups bonded to a molecule and their molar mass (notice how the global properties of smaller graphs can be the node properties of bigger graphs); since these are continuous values, there is no need to one-hot encode them. This results in a 17 x 2 tensor. On the other hand, we can extract the total reaction Gibbs free energy from each edge (again, a continuous value), while also identifying the presence of an intermediary molecule, byproduct, or cofactor (e.g. the presence of ATP or NAD+ during a reaction step) which are encoded using one-hot vectors (as seen in Figure \ref{Figure11}), therefore generating a 2D edge tensor of dimension 16 x 3. Finally, we obtain the overall standard free energy change of the reaction system and the net amount of ATP molecules produced as globals, leading to a final 1 x 2 tensor. Note that these tensors represent some information from glycolysis but not all of it. Together with the adjacency matrix (17 x 17), this graph gives us a total of 4 tensors once again (Figure \ref{Figure11}).

\FloatBarrier

\begin{figure}[!htb] \centering
  \centering
  \includegraphics[height=0.8\textheight]{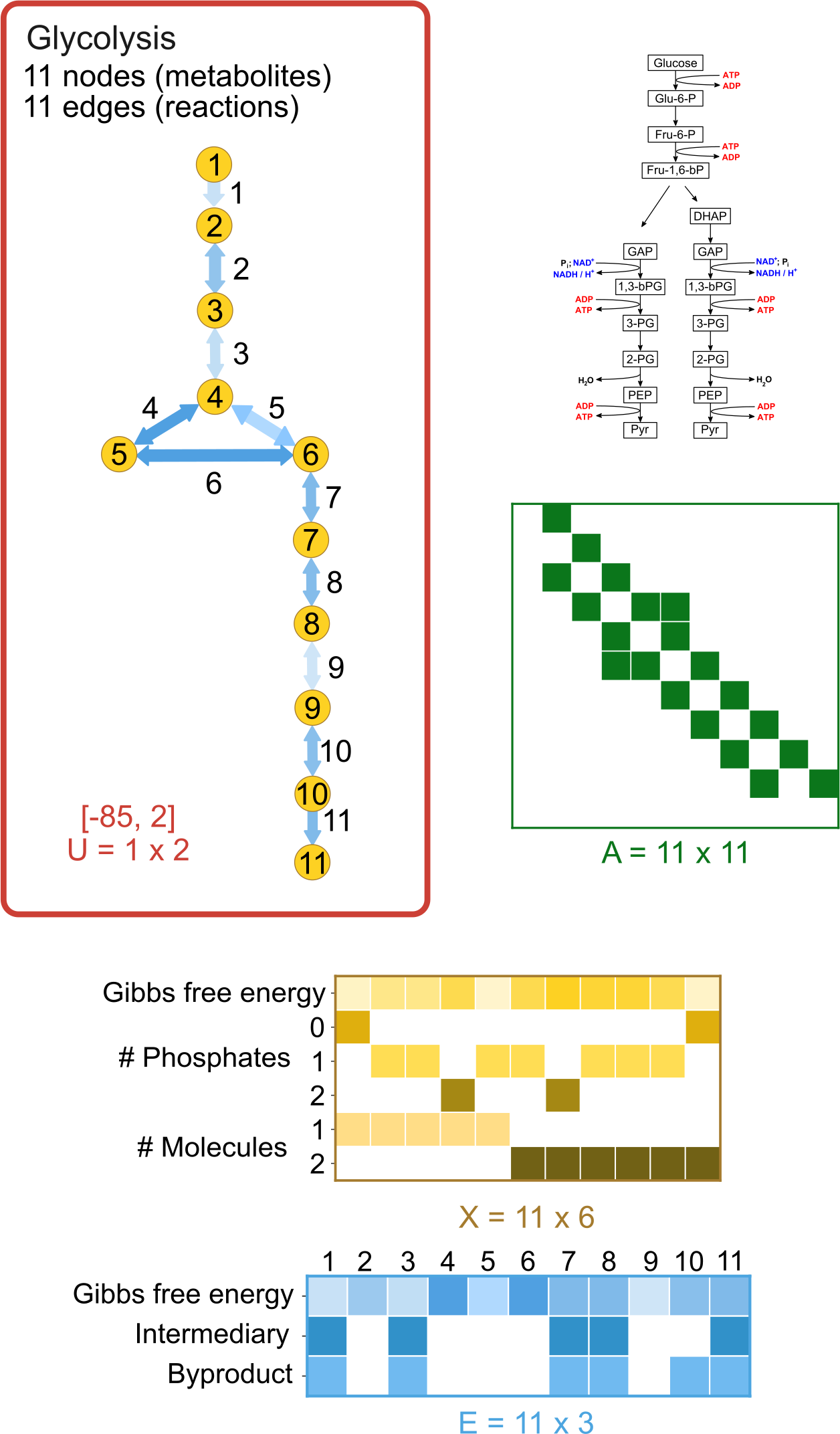}
  \mycaption{Graph tensor for glycolysis with directionality}{Showcase of adjacency matrix (A), node tensor (X) with dimensions 11 nodes x 6 features, edge tensor (E) with dimensions 11 edges x 3 features, and global tensor (U).}
  \label{Figure12}
\end{figure}

Another way to describe glycolysis as a graph is by simplifying the final 5 steps to obtain pyruvate. Instead of having separate reactions after node 4 (fructose 1,6-bisphosphate), we introduce another edge between node 5 and 6 (dihydroxyacetone phosphate and glyceraldehyde 3-phosphate respectively). With this change, we take the directionality in each reaction step into account, making a distinction between steps that are irreversible (one way) and those that are reversible reactions (go both ways). In this new abstraction (Figure \ref{Figure12}), each reversible reaction equals 2 edges instead of 1; this generates an edge tensor with dimensions 19 x 3 and a new adjacency matrix of 11 x 11. Moreover, we can extract a new node feature by identifying which nodes represent a pair of molecules, leading to a node tensor with dimension 11 x 6.

\FloatBarrier

We've seen that we can build different graphs of glycolysis by changing our focus. We can even create graphs within graphs. For example, G6P, a node in the glycolysis graph, could be expanded into its own graph. This idea of nested graphs is illustrated in the next example, where glycolysis is part of a broader chemical process.

\subsection{Building a chemical process graph: Tequila production}

Now, let's look at a real-world example of a complex chemical process: tequila production. The alcoholic fermentation stage, in particular, involves numerous biological and chemical reactions. This makes it a perfect case to demonstrate the versatility of graphs, even for representing industrial processes that are open to different interpretations.
To start, the tequila process has 3 initial steps: the preparation of the must, fermentation, and distillation. After we obtain the distilled tequila, we can age it in 3 different ways or time frames (reposado, añejo, and extra añejo), but for now let’s group all into just 1 aging step. After tequila is aged, each variety can be either blended to create flavor mixtures or filtered to remove color and produce cristalino. Once we have filtered or blended tequilas, we filter them one more time and dilute them to obtain the right alcohol \%. In addition, unaged tequila or aged tequila that is not blended or filtered can be directly diluted. In the final step, the different tequilas are bottled and sealed for selling. Based on this, we can define nodes as unit operators and edges as the flow of material between them. Let's establish this process as one with 8 unit operators (1 for each step), which makes 8 nodes, and between these nodes, we consider the different aging methods for tequila as one. The blending, aging, and filtering steps leave us with 10 edges (due to the multiple tequilas we can produce) that represent the material flow into and out of each unit operation. From the nodes, we can pull information about the time they take to operate, the costs, and energy they require, which gives us an 8 x 3 tensor. For each edge, we can extract information about the mass and volumes in inputs and outputs, obtaining a 10 x 2 tensor. As globals we could define the total energy and monetary costs of the process, as well as an efficiency parameter, resulting in a 1 x 3 1D tensor. Our fourth tensor would be the adjacency matrix with 8 x 8 dimensions.

\begin{figure}[!htb] \centering
  \includegraphics[width=1.0\columnwidth]{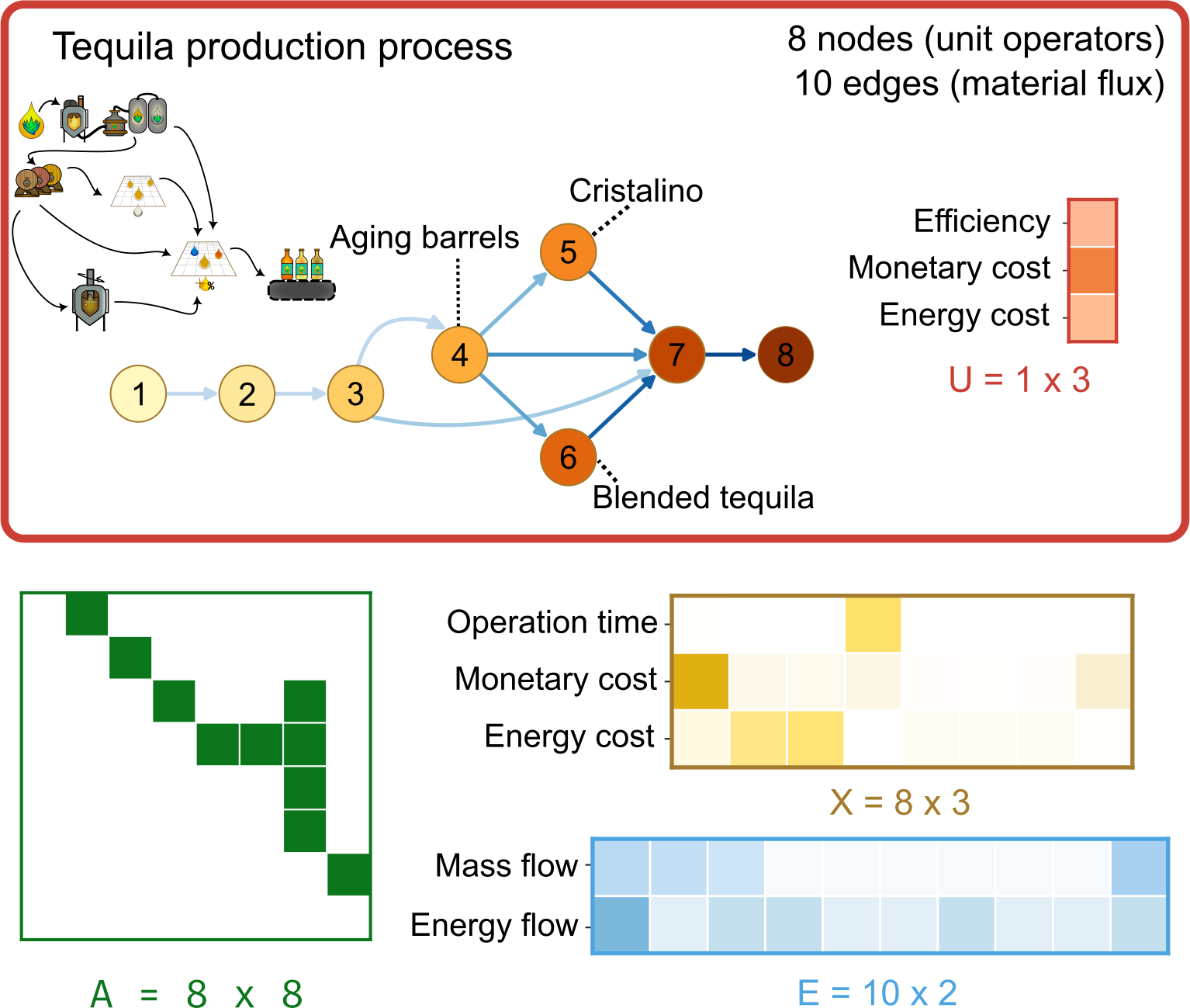}
  \mycaption{Graph tensor for the tequila process}{Showcase of adjacency matrix (A), node tensor (X) with dimensions 8 nodes x 3 features, edge tensor (E) with dimensions 10 edges x 2 features, and global tensor (U) with dimensions 1 x 3 features. 
  }\label{Figure13}
\end{figure}
\FloatBarrier

This tequila production example demonstrates how even complex systems can be represented using graphs. Other ways we can interpret tequila fermentation as a graph involve defining byproducts, recirculations, simplifying steps, or introducing hierarchical graphs (e.g. glycolysis occurs during fermentation). However, let’s try to take another approach.
We will change how we interpret the tequila production process by splitting the aging barrels node into three separate nodes representing the distinct aging processes for tequila: reposado, añejo, and extra añejo. This modification adds two nodes to the graph, updating the node tensor shape to 10 x 3.
Since each of these tequila types can be transformed into cristalino through the filtration process and are also used for blending to develop new flavors, the three new nodes are connected to both the filtration and blending nodes. Consequently, the edge tensor is updated to 18 x 2, and the adjacency matrix now has dimensions of 10 x 10.

\begin{figure}[!htb] \centering
  \includegraphics[width=1.0\columnwidth]{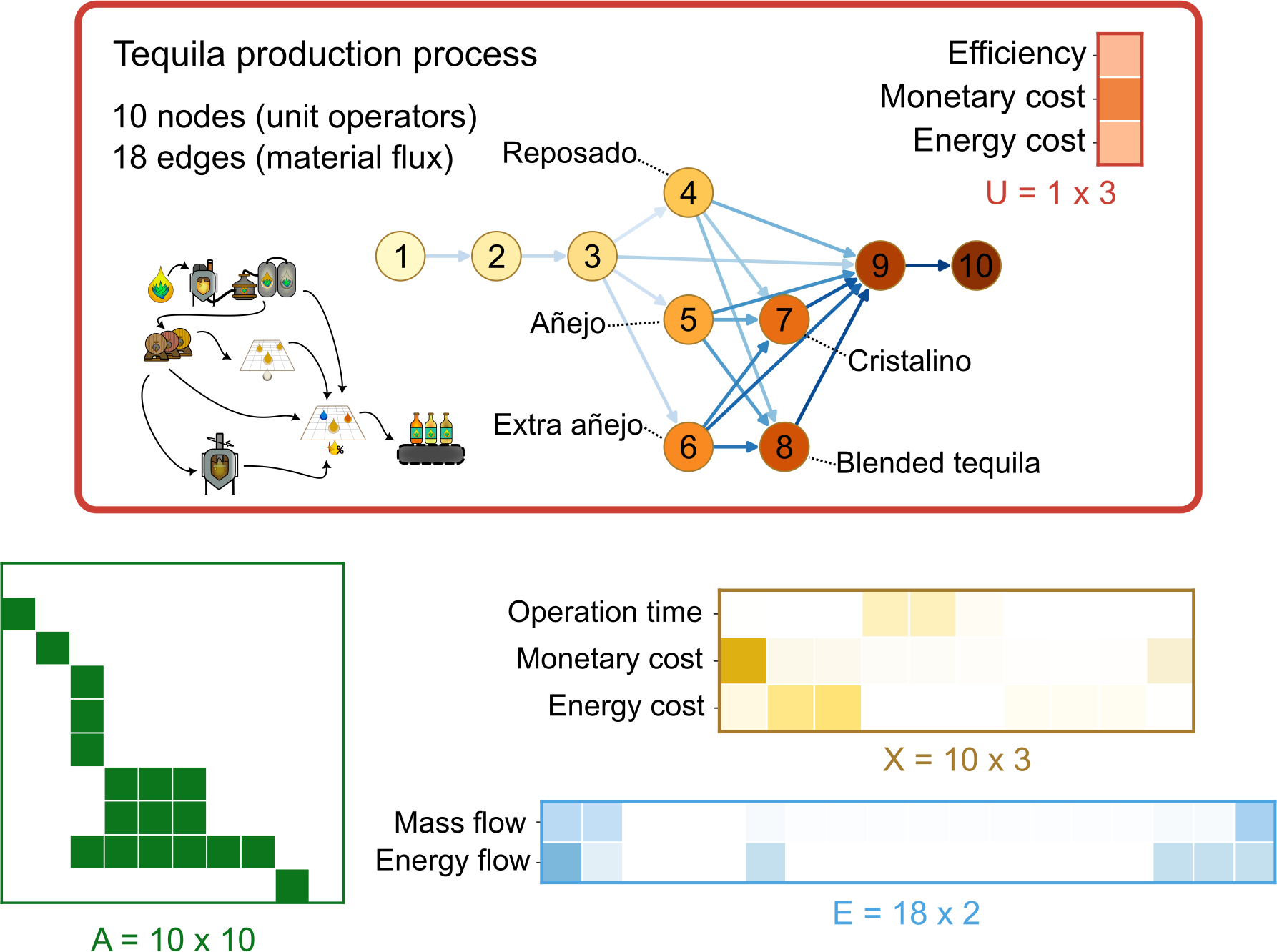}
  \mycaption{Graph tensor for the tequila process with aging split into different unit operators}{Showcase of adjacency matrix (A), node tensor (X) with dimensions 10 nodes x 3 features, edge tensor (E) with dimensions 18 edges x 2 features, and global tensor (U) with dimensions 1 x 3 features. 
  }\label{Figure14}
\end{figure}
\FloatBarrier

\subsection{Recap: Why use graphs over other representations?}

\begin{figure}[!htb] \centering
  \includegraphics[height=0.8\textheight]{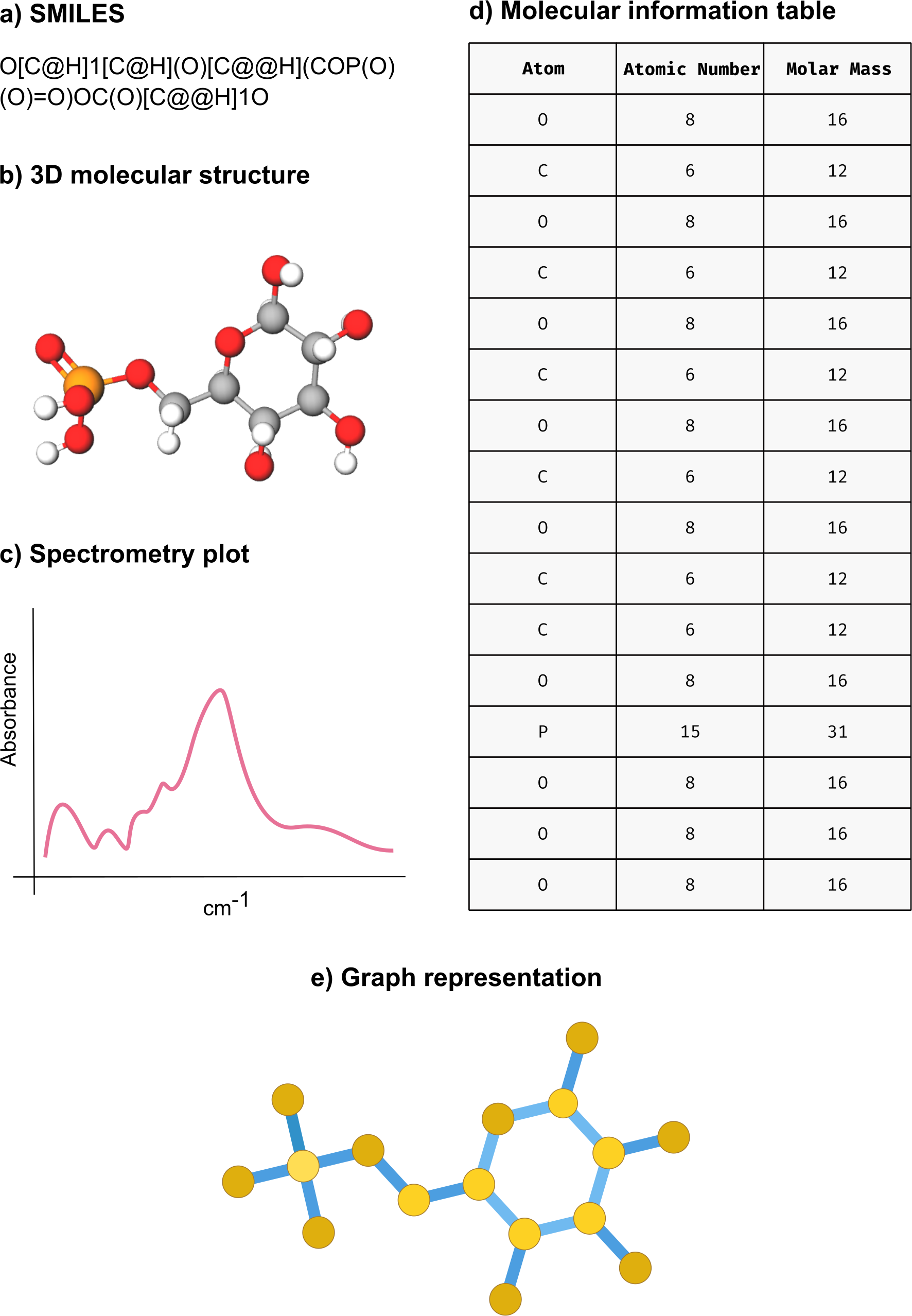}
  \mycaption{G6P representations comparison}{a) SMILES string b) Molecular image c) Spectrometry plot d) Table with molecular information e) Molecular graph.
  }\label{Figure15}
\end{figure}

We have introduced graphs, their properties, and their diverse applications in real-life scenarios. Graphs offer a practical and robust way to represent chemical systems due to their inherent flexibility and adaptability. If there is an interaction to describe, there is likely a graph that can represent it. Their components (nodes, edges, and global properties) remain fundamentally consistent across fields, allowing graphs to model diverse systems without drastic modifications. Furthermore, graphs can adapt by implementing dynamic or hierarchical features to address specific characteristics of more complex systems (time, movement, systems inside systems). These considerations allow graphs to excel in representing relational structures compared to other forms of representation, though their scalability may impose limits.
To understand why graphs excel representing chemical systems, let’s compare graphs against different forms of information. Figure \ref{Figure15} illustrates this comparison using various forms of G6P representations: a SMILES string, a table, an image, an FTIR spectrum, and a molecular graph. SMILES (Simplified Molecular Input Line Entry System) strings encode G6P’s structure linearly but fail to capture molecular symmetries. They also lack consistency, as we can have more than one SMILES representation for the same molecule. Tables can organize features like atomic mass and the number of atoms that make up G6P, but they fail to represent the relational connectivity of these atoms. To represent relationships in a table, we would require redundant information, which can be impractical or confusing in some scenarios. Images of G6P can be visually intuitive for chemists allowing them to identify the molecule, atoms, and bonds, however, it’s hard to represent complex properties like solubility or partial charges explicitly. They also depend heavily on interpretation as they offer inconsistent styling, making them impractical for computational analysis.
An FTIR spectrum provides molecular identity information, such as characteristic functional group absorption peaks or the overall molecular fingerprint, but it lacks the capacity to represent explicit atomic or bond connectivity. In contrast, molecular graphs integrate the relational and structural information missing from these other representations. They can not only describe the spatial and bonding relationships within a molecule, but also encode additional data as tensors such as atomic features and bond properties enabling computational models to process and analyze these structures efficiently.

\FloatBarrier

However, graphs are not perfect. They can be computationally expensive for large systems and require detail on their construction to ensure accurate description of systems. Nevertheless, graphs flexibility provides significantly more advantages than disadvantages for their use in chemistry. This comparison illustrates the unique capabilities of graphs and offers insight into their widespread adoption in the field.
Now that we have settled the foundations of graphs, we should ask the following questions: what kind of problems can we solve with them? What kind of limitations do they have? And most importantly, how do we actually make predictions with them?

\section{2. What kind of problems can we solve with graphs?}

We have established what a graph is and its underlying components. Now, let's examine the problems we can tackle with graph data. Specifically, we care about problems (tasks) where we want to predict an output given an input. This is a very general framing that covers a wide variety of chemical questions like molecular property prediction, reaction prediction, and molecular generation, among others. To organize these diverse applications, we introduce a task taxonomy, categorized by the graph component central to each prediction.

\subsection{Global-level tasks}

\begin{figure}[!htb] \centering
  \includegraphics[width=1.0\columnwidth]{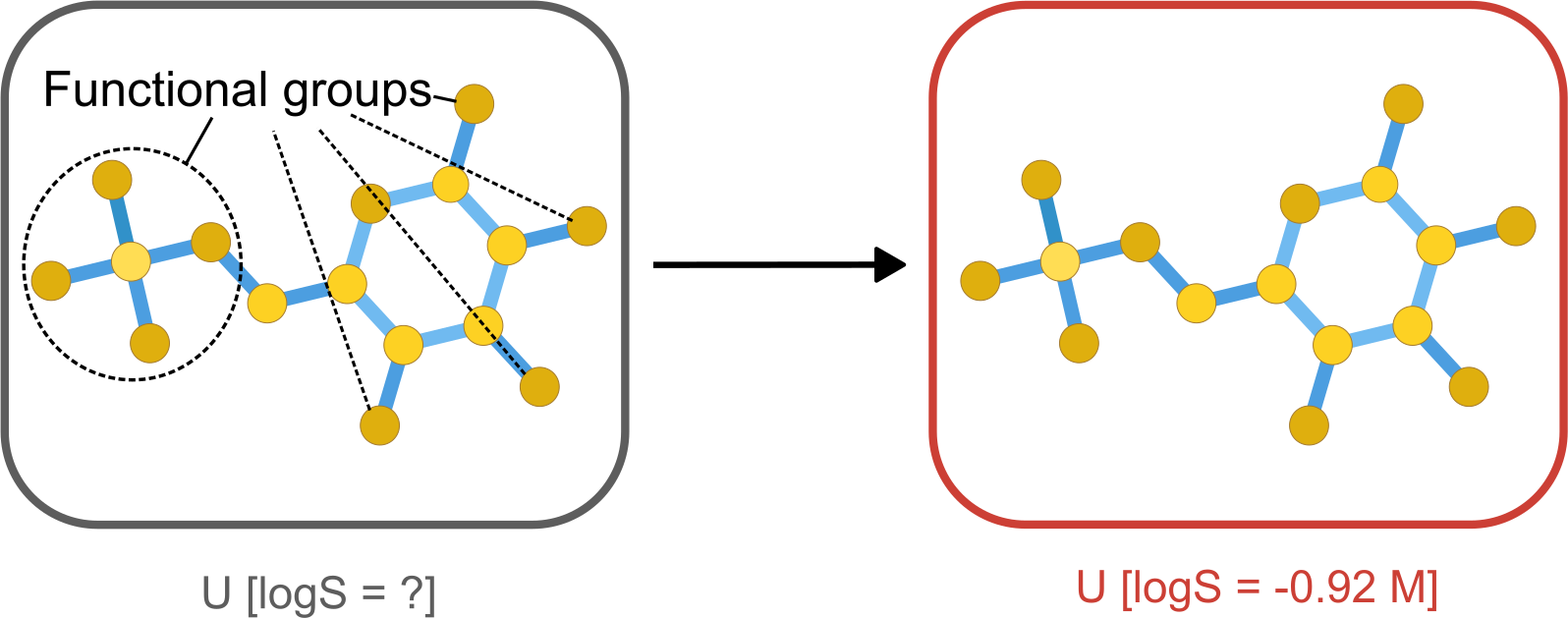}
  \mycaption{Global-level task predicting global variable U, which is the LogS value}{Based on the molecule's graph functional groups and overall structure, we estimate the expected aqueous solubility.
  }\label{Figure16}
\end{figure}

A global-level task is one that asks a question based on an entire graph. For a molecular graph like G6P, a global task can be predicting the molecule’s lipophilicity or the aqueous solubility value (logS) as seen in Figure \ref{Figure16}. We can apply a similar logic for proteins, which in essence are molecular graphs in a higher scale. Protein global-tasks involve predicting properties like activity values or their 3D structure. For a reaction pathway, one global-level task is predicting reaction dynamics (e.g. how the reaction rate changes in response to varying concentrations of reactants), another is the pathway’s yield. For an industrial graph of a chemical process, predicting the overall industrial system status determining if the system is running normally or is faulty is a global-level task.

\FloatBarrier

\subsection{Node-level tasks}

\begin{figure}[!htb] \centering
  \includegraphics[width=1.0\columnwidth]{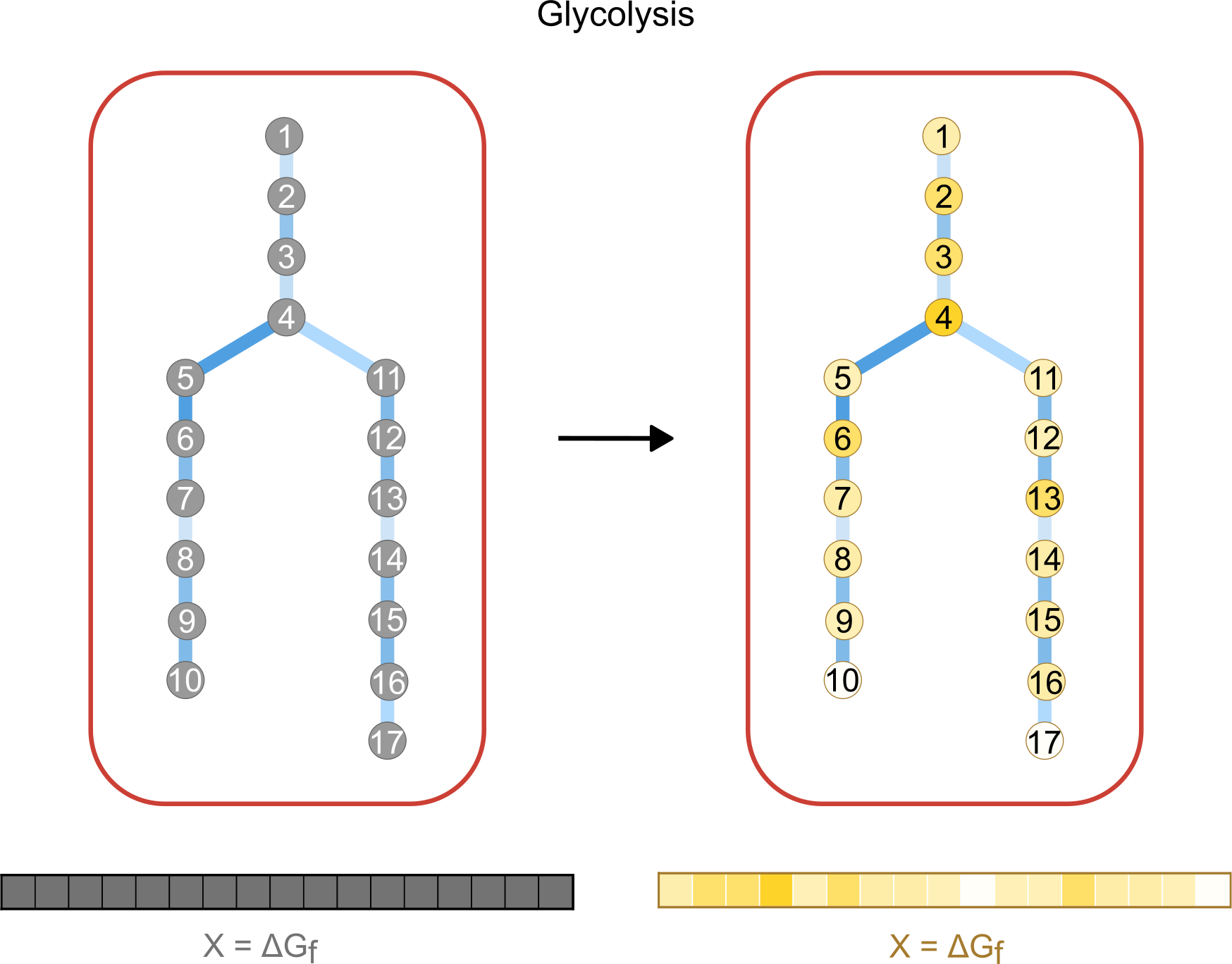}
  \mycaption{Node-level task predicting formation Gibbs free energy}{By identifying the identity of the nodes in the reaction we obtain their formation Gibbs free energy.
  }\label{Figure17}
\end{figure}

A node-level task asks a question based on the nodes of a graph. For molecular graphs, this encompasses predicting atomic binding energy, partial charges, and identifying participating atoms in reactions. Following this idea, node-level tasks for proteins determine properties over amino acid residues, such as dihedral angles or identifying which nodes are part of the active site of an enzyme. Applying this framework to reaction networks, node-level tasks are effectively employed on molecules. Consequently, we can determine formation Gibbs free energy for reactants and products (Figure \ref{Figure17}), and quantify the importance of individual molecules within reaction dynamics and product yield. In industrial graphs, node-level tasks are readily applied to unit operators to deliver predictions of key status indicators, such as operational faults.

\FloatBarrier

\subsection{Edge-level tasks}

\begin{figure}[!htb] \centering
  \includegraphics[width=1.0\columnwidth]{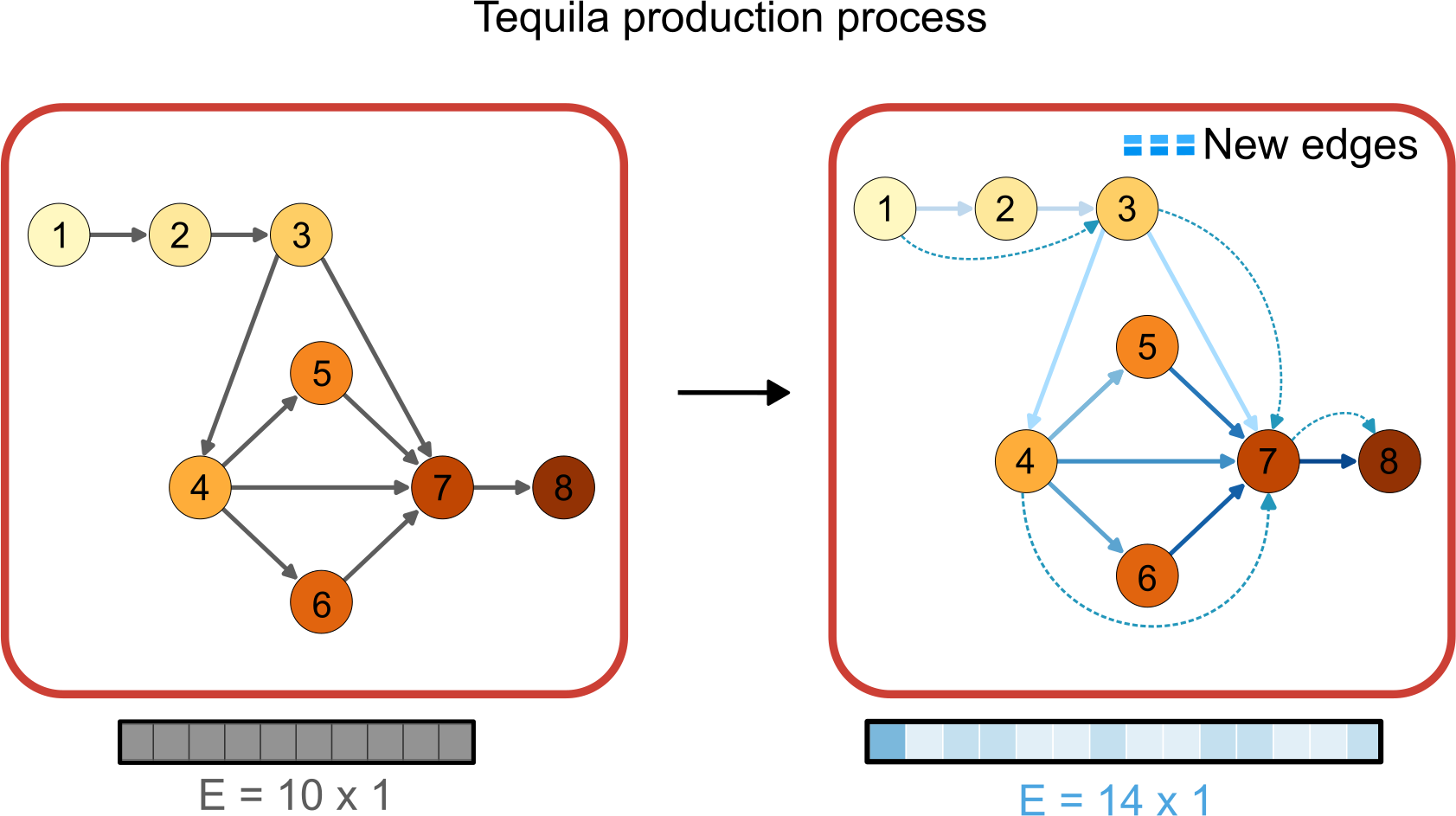}
  \mycaption{Edge-level task predicting edge properties and new edges}{New relationships between nodes and new edge information can be obtained through edge-level tasks. 
  }\label{Figure18}
\end{figure}

An edge-level task asks a question based on the edges of a graph, often a numerical property of an existing edge or inferring the presence or absence of an edge. In molecular graphs, examples include predicting bond dynamics over time and bond dissociation energy. In protein context, edge-level tasks can determine properties of peptide or hydrogen bonds, such as their distance or their presence during a reaction involving an enzyme. For reaction network graphs, the reaction Gibbs free energy of each step is an edge-level task. Finally, on industrial graphs, an edge-level task example is predicting signal dependencies between sensors of unit operators, useful to traceback failures in chemical processes (Figure \ref{Figure18}).

\FloatBarrier

Now, consider the three levels of prediction we just discussed. If we can predict node, edge, and global properties, it stands to reason that we can also predict and construct the structure of a new graph. Therefore, we could view generative modelling as a graph-level task, where instead of introducing graphs and features to predict a single node, edge, or global property, we create an entire new graph structure based on properties we desire it to have.

\subsection{Graph-level tasks}

\begin{figure}[!htb] \centering
  \includegraphics[width=0.9\columnwidth]{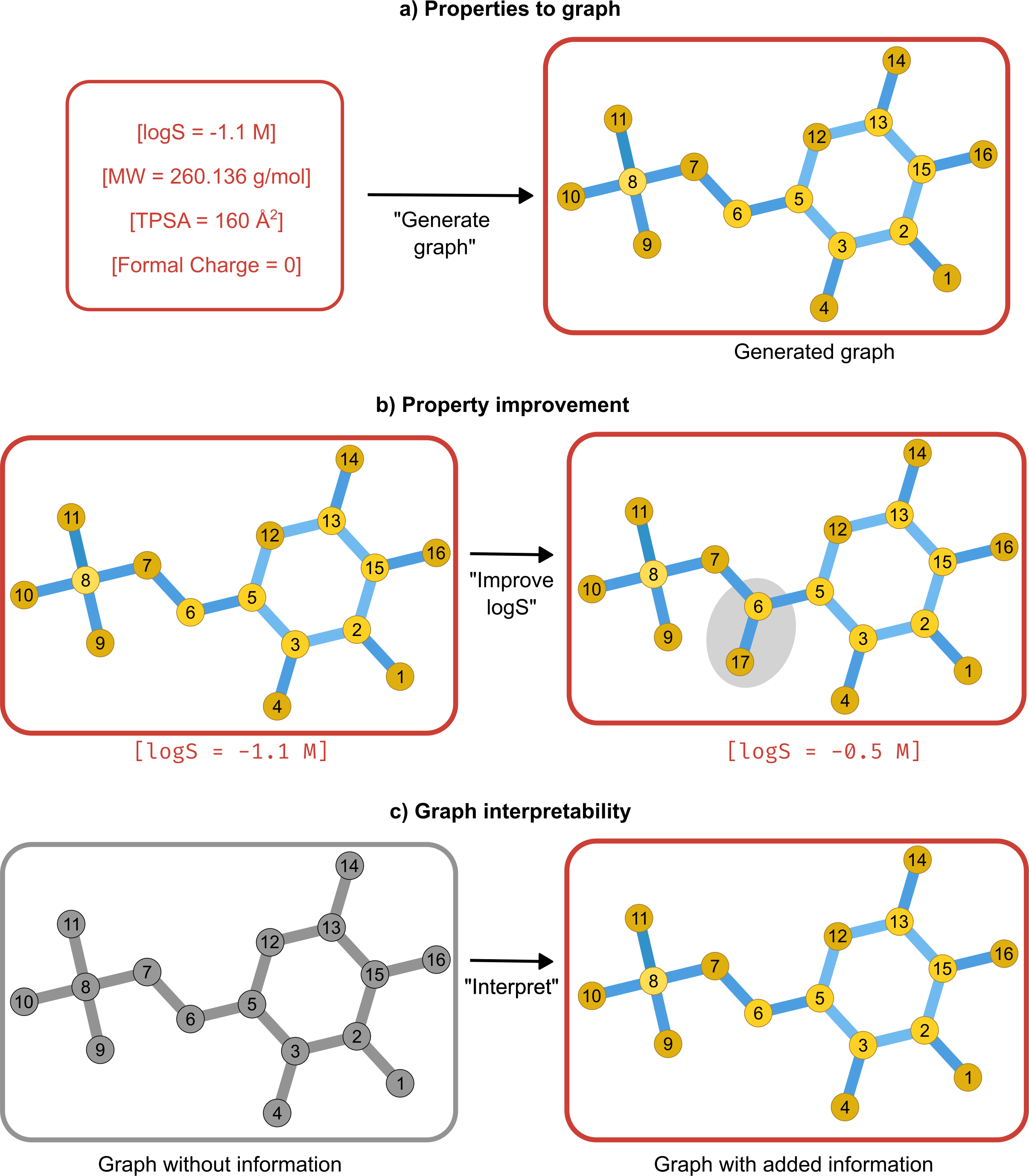}
  \mycaption{Different types of graph-level tasks}{a) Generating a graph from properties b) Improving a property by changing the graph c) Interpreting a graph.
  }\label{Figure19}
\end{figure}

A graph-level task asks a question that can be answered with a new graph. This means predicting new nodes, edges, and globals. This level of task differs from the others as it aims to predict (generate) a completely new graph structure instead of a specific feature (inverse design), it takes as input the desired properties we want the new graph to have \cite{3}.
By learning from node, edge, and global data, we can not only predict molecular properties but also generate new molecular graphs with specific characteristics (Figure \ref{Figure19}a). Depending on the application, the connectivity of the graph might remain the same or change. In interpretability tasks, for example, we may generate a heatmap over all graph components, providing a numerical rating that represents each component’s contribution to a prediction. In contrast, in generative modeling, we might modify a graph to optimize particular properties of interest.
For example, if we work with graphs containing information about mechanical features, new molecular graphs can be generated with enhanced mechanical properties such as increased tensile strength or density \cite{3,4} (Figure \ref{Figure19}b). These predicted molecular structures can then be synthesized and tested for validation. Similarly, graph-level tasks can be applied to chemical reactions, where reaction route parameters can be optimized, or entirely new reaction pathways can be generated. By coupling this approach with inverse design, a predicted molecular structure with desired properties can serve as input for another graph-level task that generates a synthesis pathway for the new molecule. Just like molecules, the generated reaction pathway can be experimentally validated \cite{4}. Furthermore, by identifying each of the predicted reaction steps, these pathways can be scaled to industrial unit operations, enabling the design of new industrial processes derived from generated chemical reactions.
As mentioned, another significant application of graph-level tasks lies in the interpretability of chemical systems and models. This approach allows for the augmentation of information available on molecules, reactions, or chemical processes. For instance, if additional information is needed about a specific molecule or reaction, a graph-level task can augment the graph representation by inferring missing or desired properties of any graph component, whether it be edge, node, or global features (Figure \ref{Figure19}c). This type of task will return the same system with additional information, which may take the form of numerical values assigned to each component or a descriptive summary of the system’s characteristics.
Inverse design approaches driven by graph representations and machine learning have the potential to fundamentally transform chemical research and development. These methods offer an efficient and accurate way to generate new molecules and processes, ultimately advancing the capabilities of molecular and chemical design.

\FloatBarrier

\newcolumntype{P}[1]{>{\raggedright\arraybackslash}p{#1}} 

\begingroup
\footnotesize                    
\setlength{\tabcolsep}{3pt}      
\renewcommand{\arraystretch}{0.9}

\begin{table}[htbp]
  \centering
  \caption{Table of task examples.}
  \label{tab:task_examples}
  \begin{tabularx}{\linewidth}{
      P{0.35\linewidth}  
      P{0.16\linewidth}  
      P{0.20\linewidth}  
      >{\centering\arraybackslash}p{0.08\linewidth} 
      P{0.07\linewidth}  
    }
    \toprule
    \textbf{Problem / Task} & \textbf{Input Data (Graphs)} & \textbf{Outputs} & \textbf{Task Type} & \textbf{Reference} \\
    \midrule
    aqueous solubility prediction & molecular & logS & global & \cite{5} \\
    lipophilicity & molecular & molecule lipophilicity & global & \cite{5} \\
    chemical reaction yield prediction & reaction networks & reaction yield & global & \cite{6} \\
    reaction dynamics & reaction networks & reaction robustness & global & \cite{7} \\
    system assessment (faulty/functional) & industrial & overall industrial system status & global & \cite{8} \\
    gene essentiality prediction & reaction networks & growth rate & global & \cite{9} \\
    atomic partial charge & molecular & partial charge values & node & \cite{10} \\
    formation Gibbs free energy & reaction & energy values & node & \cite{11} \\
    identifying molecule importance & reaction networks & quantitative importance of molecules & node & \cite{12} \\
    unit operator state (faulty/functional) & industrial & faulty unit operator detection & node & \cite{8} \\
    bond dynamics over time & molecular & updated molecule in time & edge & \cite{13,14} \\
    reaction Gibbs free energy & reaction & energy values & edge & \cite{11} \\
    bond dissociation energy & molecular & dissociation energy value & edge & \cite{14} \\
    reaction prediction & reaction networks & possible reactions between molecules & edge & \cite{12} \\
    unit operator signaling & industrial & signal status between unit operators & edge & \cite{8} \\
    molecule generation & mixed (constraint, another molecule) & molecular graph & graph & \cite{4} \\
    reaction prediction & mixed (constraint, another reaction) & reaction graph & graph & \cite{3,4} \\
    \bottomrule
  \end{tabularx}
\end{table}
\endgroup

\section{3. Expanded graph applications}

We have focused on four primary types of graphs in chemistry so far, but this list is only a representative sample; it’s not exhaustive. As we mentioned, graph data in chemistry can capture a wide variety of phenomena depending on our goals and interpretations. What other abstractions can we create with graphs? In the upcoming sections, we will explore additional types of graphs and applications, further illustrating the flexibility and power of graph representations in the chemical context.

\subsection{Reactions as graphs}

We know we can describe metabolic reactions as a graph, in a perspective this can be considered as a reaction graph, however, what about the simple step chemical reactions? These reactions can also be modeled as graphs, allowing us to represent how molecules transform into products. We can mark each of the atoms in reactant molecules to match their corresponding position in the product molecule. For example, studies like \cite{15} mapped reaction graph atoms in order to have a better understanding of reaction mechanisms and predict reaction products.

\begin{figure}[!htb] \centering
  \includegraphics[width=1.0\columnwidth]{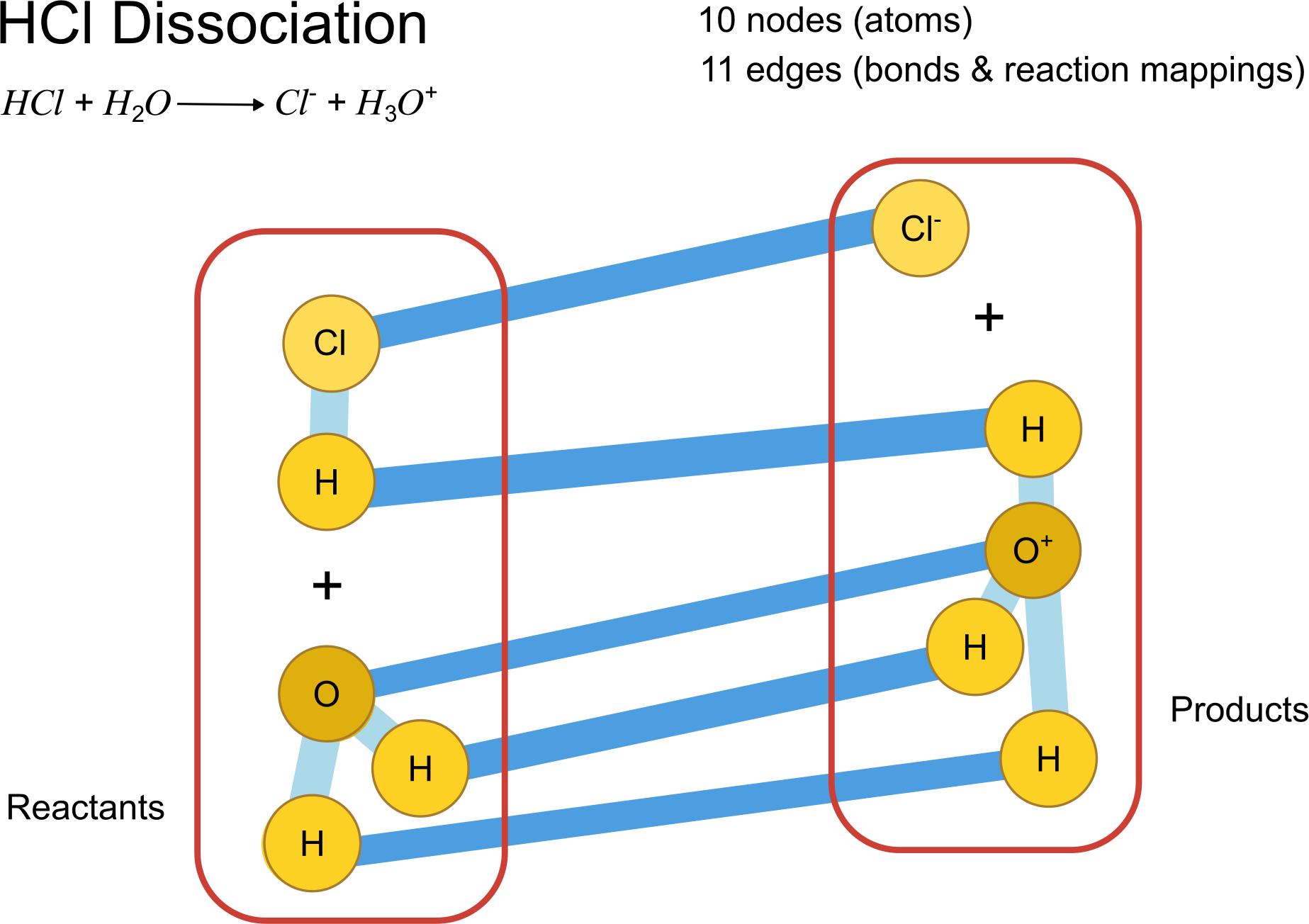}
  \mycaption{Hydrochloric acid dissociation graph, mapping reactant and product atom positions}{Reaction graph of hydrocloric acid dissociation in water, where one type of edge represents bonds between atoms and another type of edge maps the position after dissociation.
  }\label{Figure20}
\end{figure}
\FloatBarrier

\subsection{3D graphs}

There are graphs that contain 3D spatial information about the system they represent. When we work with molecules this 3D information most commonly refers to the spatial properties of atoms and bonds (coordinates derived by angles of bonds). To represent this in a graph, the 3D coordinates of atoms are typically encoded as node features in the x, y, and z dimensions. This approach supports equivariant 3D transformations, ensuring that the model captures the intrinsic 3D properties of a molecule regardless of any rotation, translation, or reflection. Therefore allowing us to represent spatial information about molecules like stereochemical structures, therefore, providing insight on stereochemical properties. We can also calculate features for node/edges such as bond distance, angle or steric hindrance related properties.
We can build upon our previous example of the 1L2Y protein to illustrate how 3D structural information can be leveraged to represent the interactions between residues. The spatial organization of residues in a protein structure provides critical insights into their functional and structural relationships. By considering the number of interacting residues within specific distance thresholds, such as 5 Å and 10 Å, we can quantify these interactions and observe how they influence the representation of the protein.
For instance, if we define an interaction as the presence of another residue within a cutoff distance, collecting interaction data at 5 Å captures the close-range, potentially stronger interactions, while at 10 Å, we include both close and more distant interactions. When these interaction data are represented in an adjacency matrix noticeable changes emerge. The adjacency matrix at 5 Å will typically be sparser, highlighting only the closest connections, whereas the matrix at 10 Å will show a denser network of interactions, reflecting a broader range of residue relationships.

\begin{figure}[!htb] \centering
  \includegraphics[width=1.0\columnwidth]{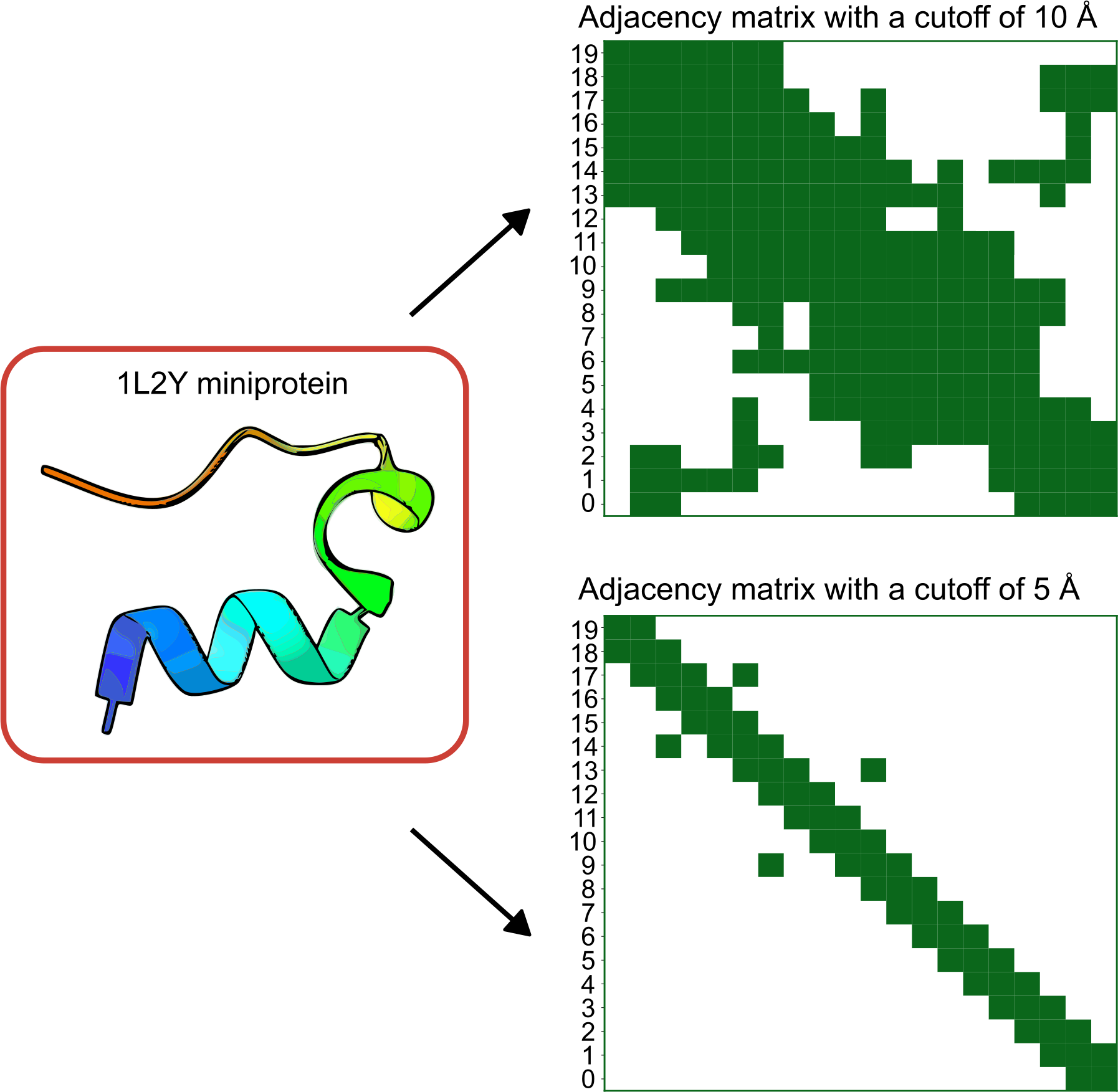}
  \mycaption{The adjacency tensor of 1L2Y protein the with a cutoff of 5 \AA{} and 10 \AA{}}{Showcasing the change from close relationship of interaction residues (sparse matrix) to a more comprehensive network of connections (dense matrix).}
  \label{Figure21}
\end{figure}
\FloatBarrier

\subsection{Time-dependant graphs}

\begin{figure}[!htb] \centering
  \includegraphics[width=1.0\columnwidth]{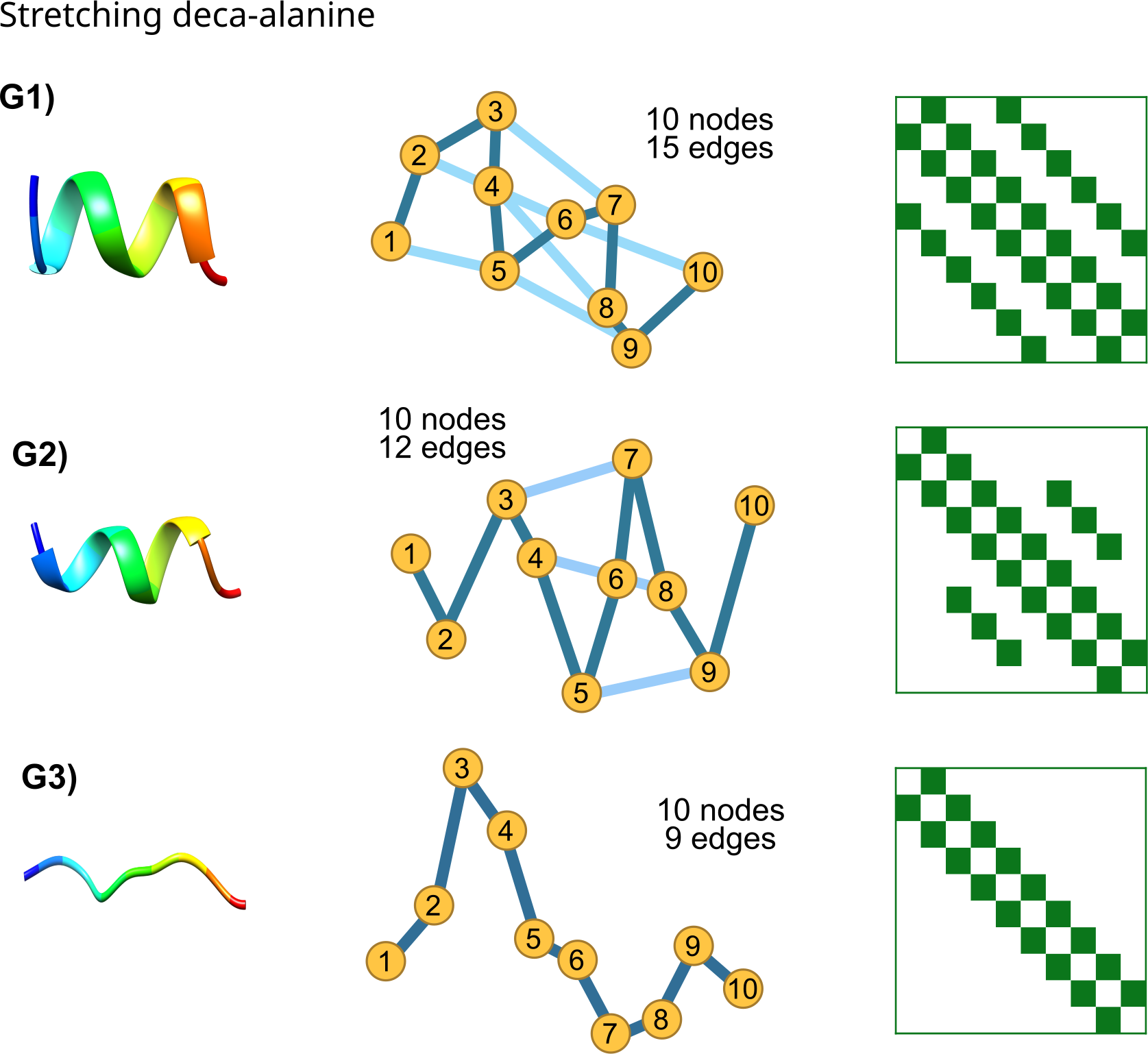}
  \mycaption{The simulation of the deca-alanine peptide stretching as a graph progression}{The peptide starts with a packed configuration (G1). There are hydrogen bonds between nonconsecutive residues. As the peptide stretches, some hydrogen bonds break, leading to changes in the number of edges and thus altering the graph representation of the system (G2), until all of these bonds are broken (G3). These changes in connectivity can be observed in the corresponding adjacency matrices of each graph.
  }\label{Figure22}
\end{figure}

To fully understand the behavior of some systems, time is an important variable in consideration; in terms of graphs we can represent different time instances of chemical systems in multiple ways, like associating time intervals to nodes and edges, portraying multiple snapshots of the graph across time, or connecting nodes through time with edges. However, it is usually easier to just represent time as a series of discrete time steps with multiple graphs. This type of representation is often used to describe molecular dynamics or reaction kinetics.
Consider the example of the deca-alanine peptide stretching process. This peptide is often used in molecular dynamics simulation tutorials as an illustrative case \cite{16,17}. At the start of the simulation, the deca-alanine peptide starts in a packed configuration, where the 10 residues are close to one another, allowing hydrogen bonds to form between nonconsecutive residues. As the simulation progresses, the residues separate and the hydrogen bonds gradually break.
We can represent these structural changes during the simulation as a sequence of graphs (Figure \ref{Figure22}).
In the first graph (G1) we represent the peptide with 10 nodes, each corresponding to one of the peptide’s residues. The edges in this graph represent the connections between the residues. When the peptide is packed together, the graph has 10 nodes with 14 edges: 9 from peptide bonds and 6 from hydrogen bonds. 
If we inspect the simulation in another point (G2), as the peptide stretches, 4 of the 6 hydrogen bonds are broken. We can apply this same logic when the peptide completely stretches (G3) when further hydrogen bonds break and the edges reduce to only the 9 peptide bonds between the amino acids.

\FloatBarrier

\subsection{Crystal Structure Graphs}

At first glance, crystal graph representations may seem similar to molecular graphs, but they differ in several key ways. First, crystals are solid structures characterized by continuous, repetitive patterns of atoms or molecules. This periodicity extends infinitely in all directions, so crystals are typically described by their smallest repeating unit: the unit cell. This unit cell can be converted into a graph, which is often used in material property prediction tasks. The composition, arrangement, and size of these unit cell graphs play an important role in modeling such systems.

\begin{figure}[!htb] \centering
  \includegraphics[width=1.0\columnwidth]{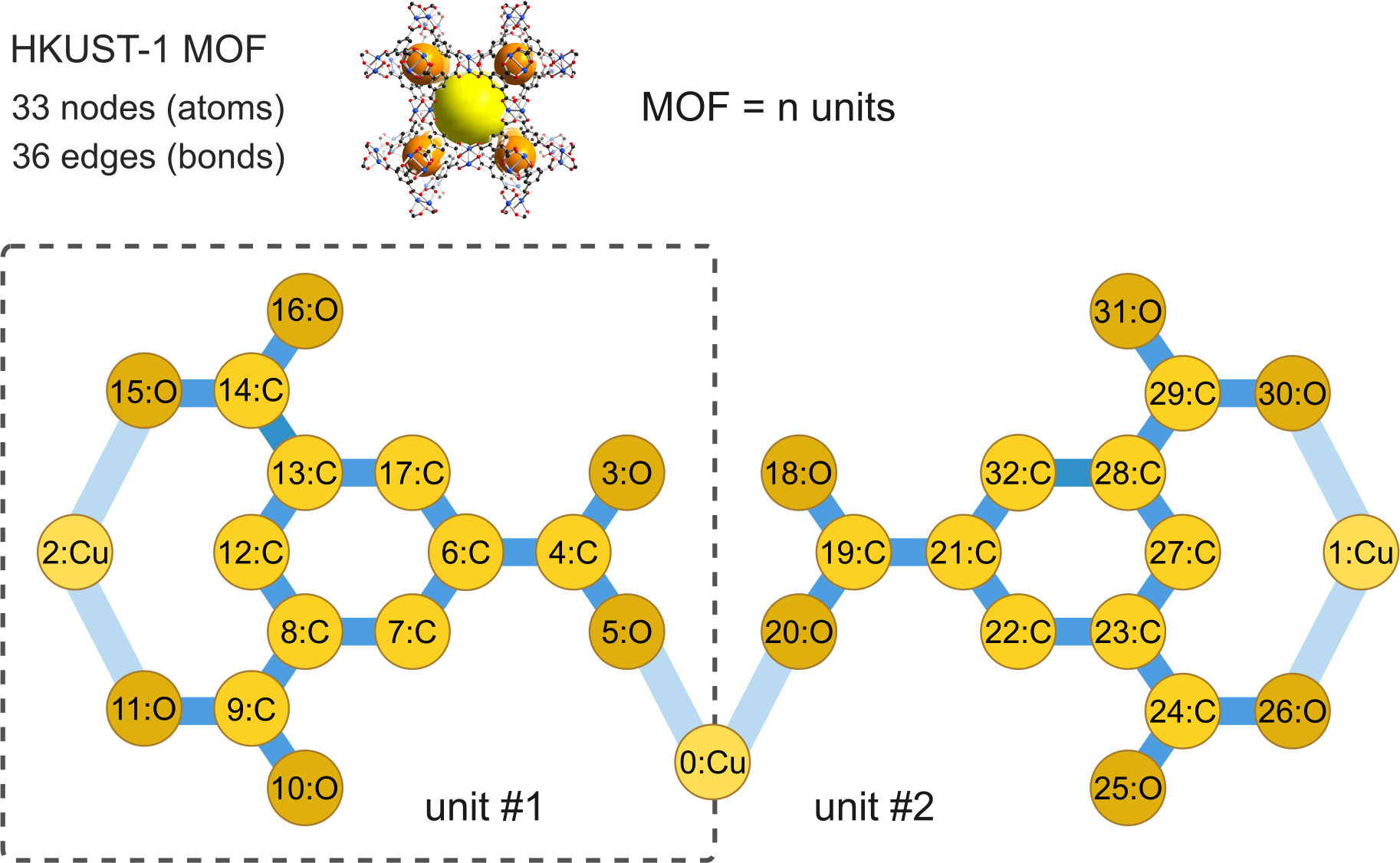}
  \mycaption{Representation of the HKUST-1 MOF, with 2 units showcased}{The spheres in the crystal structure represent pockets that can be utilized within the MOF. The dashed line box highlights an individual structural unit of the HKUST-1 MOF, which repeats to form the extended network.
  }\label{Figure23}
\end{figure}

Similarly, metal-organic frameworks (MOFs) exhibit periodic chemical structures and can also be represented as graphs. In this case, one may choose to build the graph at either the atomic level or at the level of larger building blocks—such as unit cells or repeating motifs. These MOF graphs may include multiple node and edge types, reflecting the chemical diversity of their components, which often include cycles, organic functional groups, metal centers, and small molecules, as depicted in Figure \ref{Figure23} \cite{18}.

\FloatBarrier

\subsection{Protein interaction networks}

When studying protein functions, representing them as graphs is helpful because of the complexity of their interaction networks. In these graphs, proteins are represented as nodes, their interactions as edges, and global features may include information such as organism, tissue, or cell type, allowing us to better visualize these large interaction networks\cite{19}. This approach is particularly valuable for understanding complex systems such as the protein interaction network in organisms like Mycobacterium tuberculosis.
One study\cite{20} predicted 911 protein–protein interactions in Mycobacterium tuberculosis. These interactions are fundamental to cellular processes, including signal transduction, metabolic pathways, and structural assembly. To effectively model protein–protein interaction (PPI) networks, it is essential to describe both the nodes (proteins) and the edges (interactions) in terms of their biochemical and structural properties. Node features may include sequence data, secondary structure, 3D conformation, or functional annotations. Edge features may describe interaction kinetics or binding energies between protein pairs. Additionally, global features may capture properties of the proteome of origin, providing context for the entire network.

\begin{figure}[!htb] \centering
  \includegraphics[width=0.7\columnwidth]{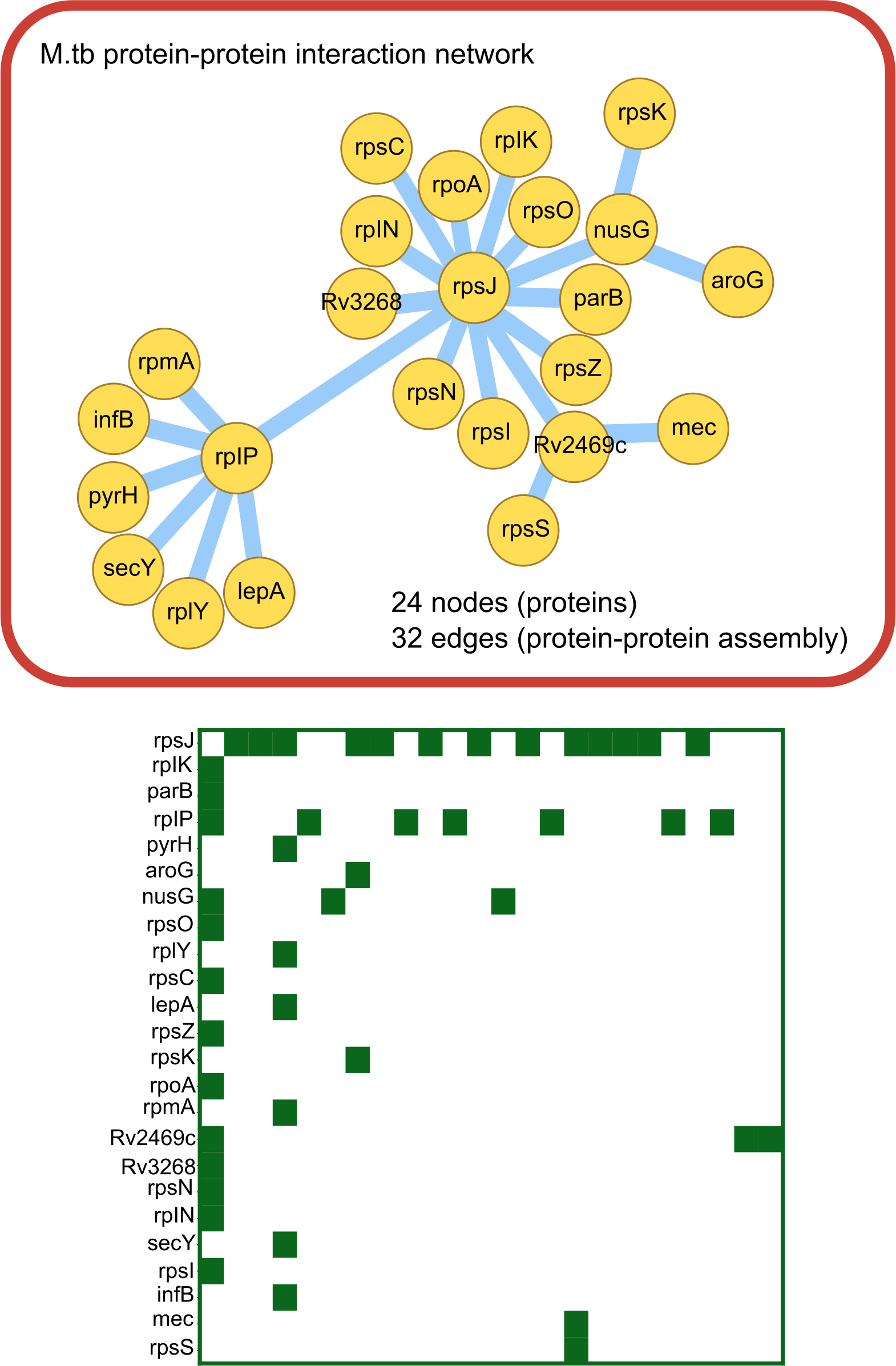}
  \mycaption{A simplified predicted PPI interaction network of 24 proteins from \textit{Mycobacterium tuberculosis}}{Showcasing protein-protein interactions among them.
  }\label{Figure24}
\end{figure}
\FloatBarrier

\subsection{Drug-target interaction networks}

\begin{figure}[!htb] \centering
  \includegraphics[height=0.8\textheight]{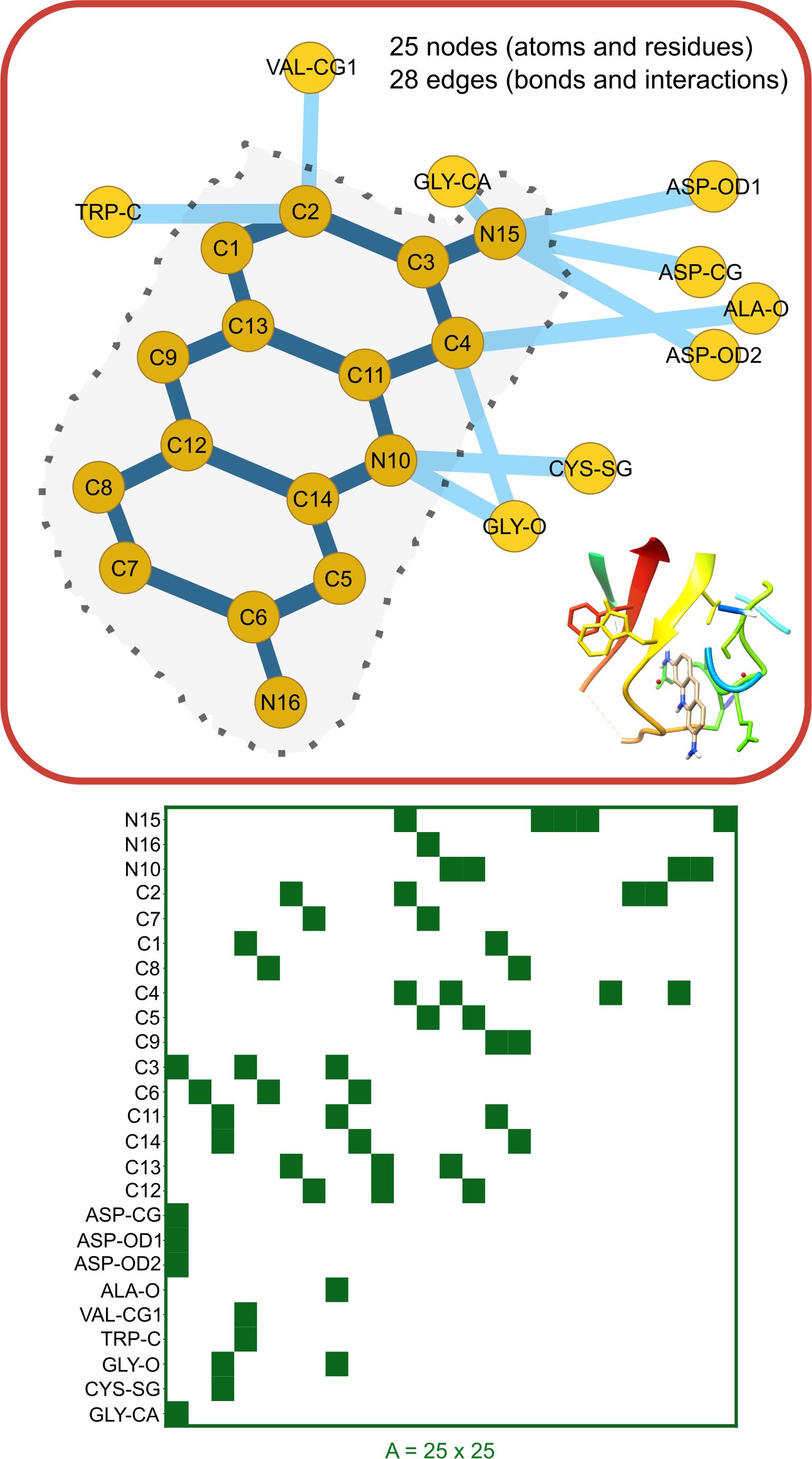}
  \mycaption{The graph representation of the binding pocket of the proflavine molecule and the thrombin protein}{Showcasing the regions where the molecule and the protein interact. The dashed line zone encloses the nodes and edges corresponding to the molecule while the other nodes and edges
  represent the enzyme.}\label{Figure25}
\end{figure}

Another example of where chemistry and biology converge is the interactions drugs have with living organisms. This mirrors protein interactions, with drug relationships often exhibiting similar complexity, scale, and involving numerous interacting components. In this case, nodes represent drugs or receptor molecules within a cell, and edges capture their interactions (e.g., a drug binding to or modulating a receptor)\cite{21}.
This interaction is exemplified by the binding between the proflavine molecule and the alpha-thrombin protein. Proflavine is commonly used to study the interactions between inhibitors and substrates within a specific protein pocket. These insights support the development of novel antithrombotic drugs\cite{22}. While the primary interaction involves a a glycine residue engaging with a nitrogen atom, and other atoms also contribute significantly to the binding of the molecule.
In the binding pocket (Figure \ref{Figure25}), two types of atoms contribute to the interaction between the molecule and the protein: atoms from both the proflavine molecule and the protein. The representation in Figure \ref{Figure25} highlights that this interaction can be modeled as a heterogeneous graph, featuring two distinct types of nodes. Similarly, the graph includes two types of edges: those connecting atoms within the molecule and those linking them to protein atoms.

\FloatBarrier

\subsection{Metabolic networks}

\begin{figure}[!htb] \centering
  \includegraphics[width=.8\columnwidth]{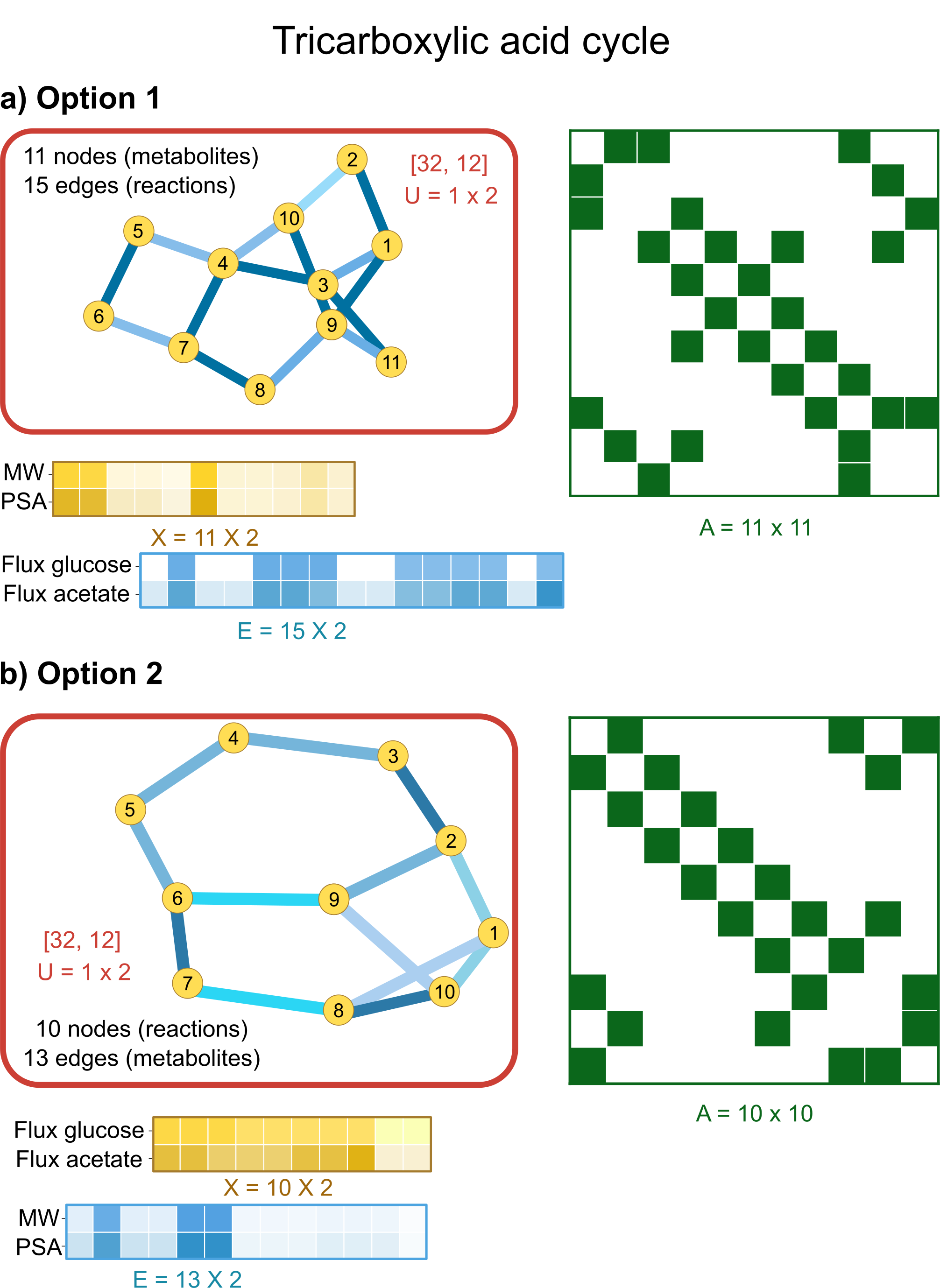}
  \mycaption{Tricarboxylic acid cycle in \textit{E. coli} graph representations}{a) This option for representing the TCA cycle in a graph considers the metabolites as nodes (1: acetyl-coA, 2: coA, 3: citrate, 4: isocitrate, 5: alpha-ketoglutarate, 6: succinyl-coA, 7: succinate, 8: fumarate, 9: malate, 10: glyoxylate, 11: oxaloacetate) and enzymatic reactions as edges b) This option flips the definitions of the nodes and edges, and defines the enzymatic reactions as nodes (1: CS, 2: ACN, 3: ICD, 4: KDH, 5: ScAS, 6: SDH, 7: FUM, 8: MDH, 9: ICL, 10: MS).
  }\label{Figure26}
\end{figure}

One interesting example in biochemical systems is representing metabolic networks and their dynamic nature in graph representation. The previously introduced glycolysis network is an example of this (Figure \ref{Figure26}), but it can be taken one step further by representing metabolites as nodes, while the reaction fluxes connecting them serve as edges, allowing for the analysis of topological properties such as network connectivity, modularity, and centrality. Another graph representation that is used to represent genome-scale metabolic networks is one that uses the nodes to represent enzymatic reactions, while the edges represent metabolite mass flow between reactions they’re involved in. Directed edges can be used to represent the flux direction and also expanded to integrate genomic and transcriptomic data, which could provide insight to metabolic flux distributions and even the effects of perturbations on the overall metabolic system \cite{23}. Both of these graph representations are presented in Figure \ref{Figure26} for the citric acid cycle in \textit{E. coli} \cite{24}. We see in panel A the first graph representation which depicts metabolites as nodes (11 different nodes) and the reactions (or enzymes) as edges (15 different edges). Panel B, illustrates the second graph representation for the citric acid cycle where the reactions/enzymes serve as nodes instead (10 nodes), while the metabolites serve as edges (13). We can see from these two representations that although they depict the same metabolic network with the same metabolite and reaction features, they have different graph structures that emphasize different aspects of the system.

\FloatBarrier

\subsection{Knowledge graphs}

Our final examples are knowledge graphs. While other chemical graphs represent relationships within a chemical system (such as bonds or connectivity), knowledge graphs represent relationships between different types of chemical information. They organize data on molecules, reactions, catalysts, and conditions into a comprehensive network that standardizes and facilitates access to chemical knowledge.
One key application of knowledge graphs is organizing and navigating domain-specific concepts through taxonomic trees. A taxonomic tree is itself a graph that structures information from broad structural classes down to fine-grained distinctions based on functional groups, properties, or biological activity. For example, a molecule can be classified as belonging to a "heterocyclic compound" class, further refined into "pyridine derivatives" and then annotated with specific properties and functionalities at each level of the taxonomy.
To illustrate, consider ChemOnt (Chemical Entities of Biological Interest Ontology). ChemOnt is designed as a hierarchical classification system for chemical entities, including molecules, reactions, and processes. It employs "is-a" relationships to build a taxonomic tree that precisely categorizes molecules from broad classes to specific instances. By integrating ChemOnt into knowledge graphs, researchers can leverage its structured vocabulary to ensure consistent, semantically rich annotations, thereby facilitating knowledge discovery and automated reasoning across chemical sets of data \cite{25}.

\begin{figure}[!htb] \centering
  \includegraphics[width=1.0\columnwidth]{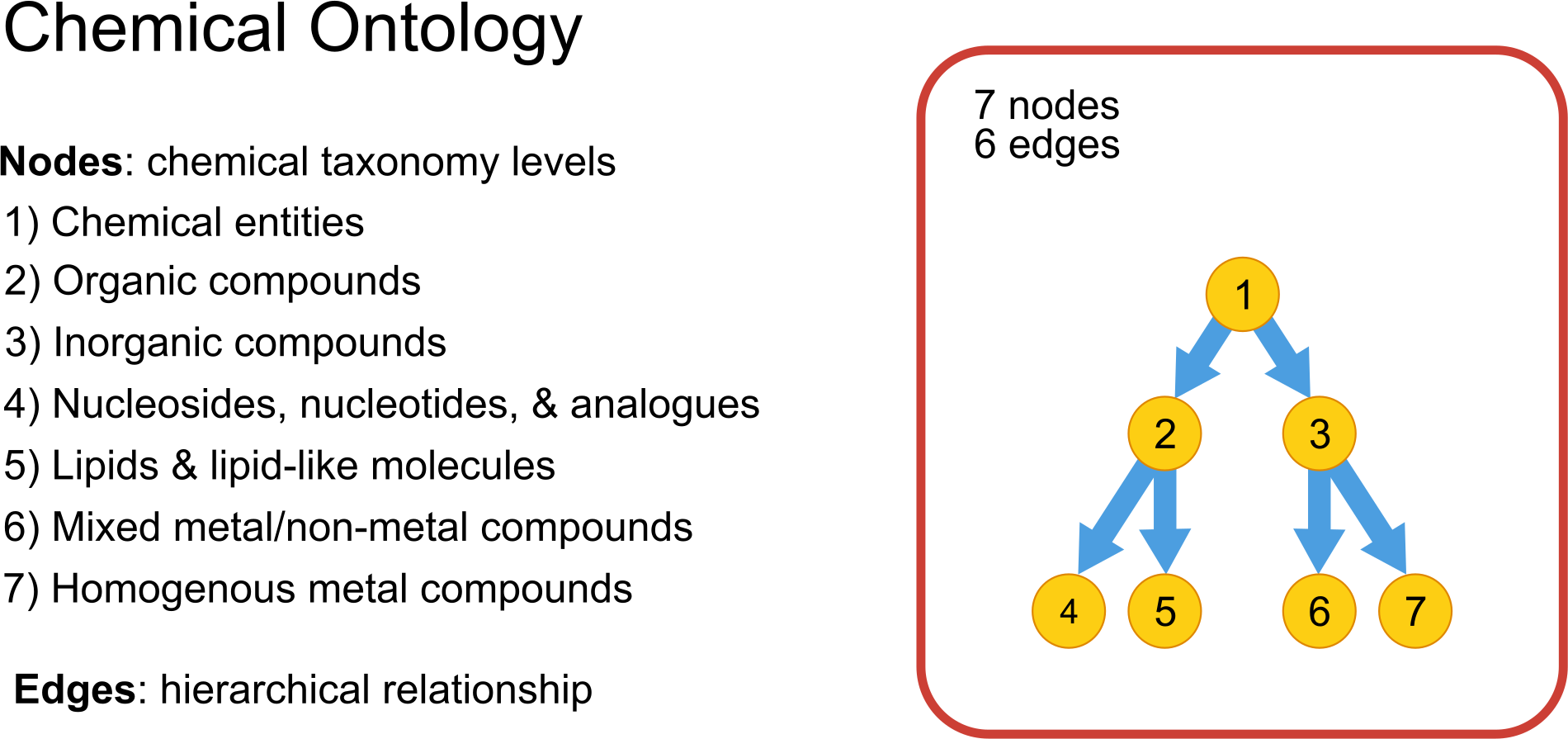}
  \mycaption{Chemical ontology classification system graph}{In this knowledge graph representation of the chemical ontology classification system, nodes correspond to chemical taxonomy levels, while edges represent hierarchical relationships.
  }\label{Figure27}
\end{figure}
\FloatBarrier

\section{4. Learning algorithms on chemical graphs}

In the previous sections, we explored how chemical systems can be represented as graphs and how diverse chemical problems can be formulated as graph-based prediction tasks. The versatility of graphs allows us to address a wide range of chemical phenomena, as demonstrated by the examples we've covered. In this chapter, we turn our attention to the computational methods used to solve these tasks. Specifically, we'll examine how machine learning (ML) enables us to make predictions and learn from graph-structured data. The fields of machine learning and more broadly, artificial intelligence, provide the tools and techniques to answer the key question: how can we leverage the information encoded in chemical graphs to gain new insights and solve challenging problems?

\subsection{Machine Intelligence - Learning Algorithms}

\begin{figure}[!htb] \centering
  \includegraphics[width=1.0\columnwidth]{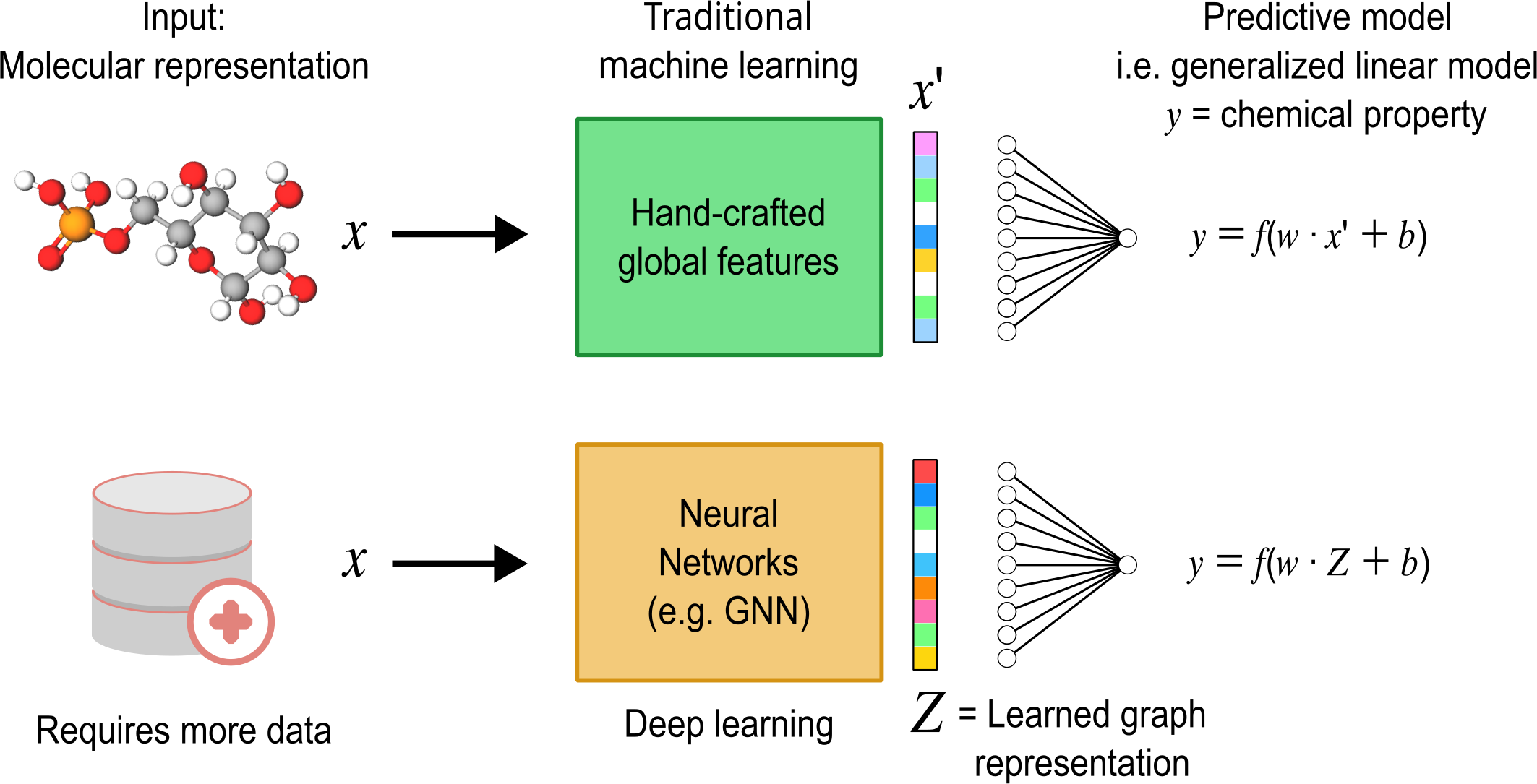}
  \mycaption{Machine learning vs deep learning framework comparison ($f$ = predictive model)}{Comparison between traditional machine learning and deep learning for chemical property prediction \it{"y"}. Traditional approaches rely on hand-crafted global features, while deep learning methods (e.g., GNNs) learn representations directly from data.
  }\label{Figure28}
\end{figure}

In machine learning, a learning algorithm improves its performance on a task (T) by learning from data (E), as measured by a performance metric (P) \cite{26}. For chemical graphs, T might be predicting properties or designing molecules, P could be prediction accuracy, and E is all the graph data. This chapter first briefly visits traditional machine learning (ML) algorithms, highlighting how researchers have historically approached the crucial step of feature engineering, manually crafting relevant features from graphs. We then transition to deep learning (DL), specifically graph neural networks (GNNs). Unlike traditional ML, GNNs automate feature extraction, learning to represent nodes, edges, and entire graphs. This end-to-end learning capability makes GNNs powerful tools for chemical applications, but their effectiveness often depends on substantial data availability, a key consideration in chemical research. Figure \ref{Figure28} highlights the comparison between the two approaches. Although earlier sections explored diverse graph applications, in this chapter we will focus primarily on molecular graphs.

\FloatBarrier

\subsection{Feature-Engineered Machine Learning}

\begin{figure}[!htb] \centering
  \includegraphics[width=1.0\columnwidth]{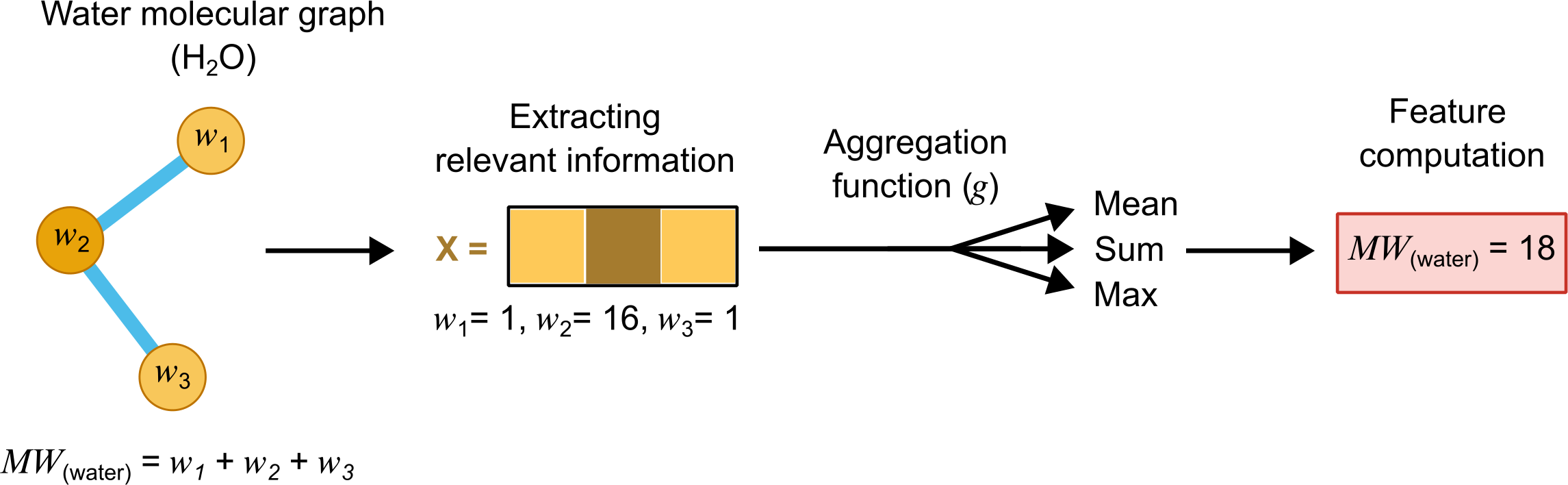}
  \mycaption{Diagram to compute molecular weight feature}{Using a water molecule as an example, node weights (atomic contributions) are extracted and aggregated using a function \it{"g"} (e.g., mean, sum, or max) to compute molecular weight.
  }\label{Figure29}
\end{figure}

Before the rise of deep learning and graph neural networks (GNNs), feature-based machine learning methods were the standard approach, and they remain valuable tools in many situations. These methods typically operate on fixed-size vector representations, requiring a crucial step called feature engineering: the manual extraction of relevant characteristics (features) from the graph data, often relying on significant domain expertise (e.g., chemical knowledge). In chemistry, these features might include molecular fingerprints, atom counts, bond types, or computed physicochemical properties like solubility. The process of transforming a chemical graph into a fixed-size vector often involves aggregation (or pooling) functions. These functions take a variable number of inputs and return a single output, crucially regardless of the order of the inputs. This property, known as permutation invariance, is essential when dealing with graph data where node order is arbitrary. Common aggregation functions include min, max, mean, sum, and variance.
Consider the calculation of molecular weight (MW) for a molecule like water ($H_{2}O$) as a simple example. First, the identities of the constituent atoms (two hydrogen atoms and one oxygen atom) are extracted. Then, the atomic masses of each atom are retrieved, forming a set of values. Finally, a sum aggregation function is applied to this set. The sum aggregation function ensures permutation invariance, so the order of addition does not affect the final MW, correctly reflecting the fact that the molecule's weight is independent of how we label the atoms (Figure \ref{Figure29}). While MW provides a single statistic, we can compute many such statistics using different aggregation functions and applying them to various graph properties, ultimately creating a multi-dimensional feature vector.

\FloatBarrier

\begin{figure}[!htb] \centering
  \includegraphics[width=1.0\columnwidth]{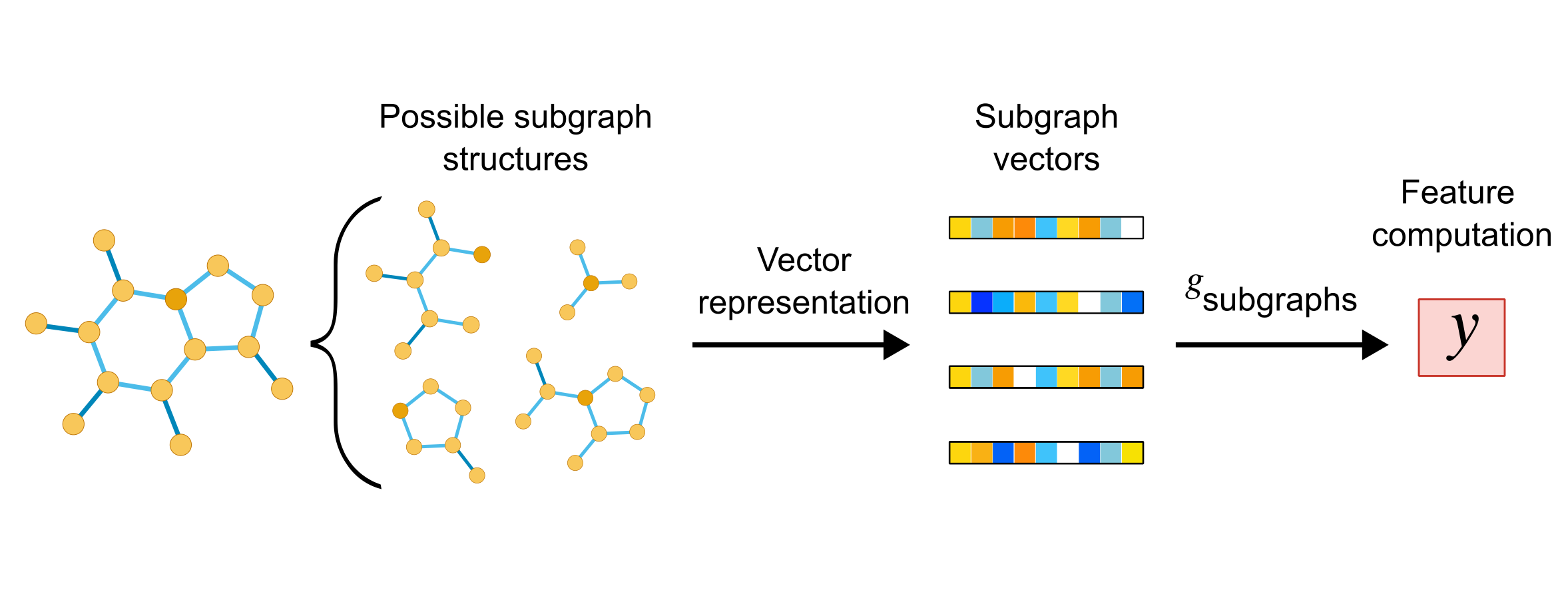}
  \mycaption{Diagram of computing subgraph features}{Possible subgraph structures are converted into vector representations, which are aggregated to compute the final feature \it{"y"}.
  }\label{Figure30}
\end{figure}

Beyond whole-graph properties, feature engineering often involves analyzing subgraphs specific structural patterns within the larger graph. These substructures can represent bond types, functional groups, ring systems (including aromatic rings), and other chemically relevant motifs. Extracting information about these substructures can involve several approaches. Some algorithms directly identify substructures based on node and edge properties. For instance, to detect aromatic rings, an algorithm might identify potential ring systems and then apply Hückel's rule to determine aromaticity. Other approaches rely on graph traversal (or graph walking) algorithms, which systematically explore the graph, collecting information about node connectivity and identifying the presence, absence, or frequency of predefined substructures as seen in Figure \ref{Figure30}. The counts or presence/absence indicators of these substructures then become features in the overall feature vector. Due to the combinatorial explosion of possible substructures, these representations are typically folded into a fixed-size vector, a process that can lead to hashing collisions.

\FloatBarrier

\subsection{Common Machine Learning algorithms}

\begin{figure}[!htb] \centering
  \includegraphics[width=1.0\columnwidth]{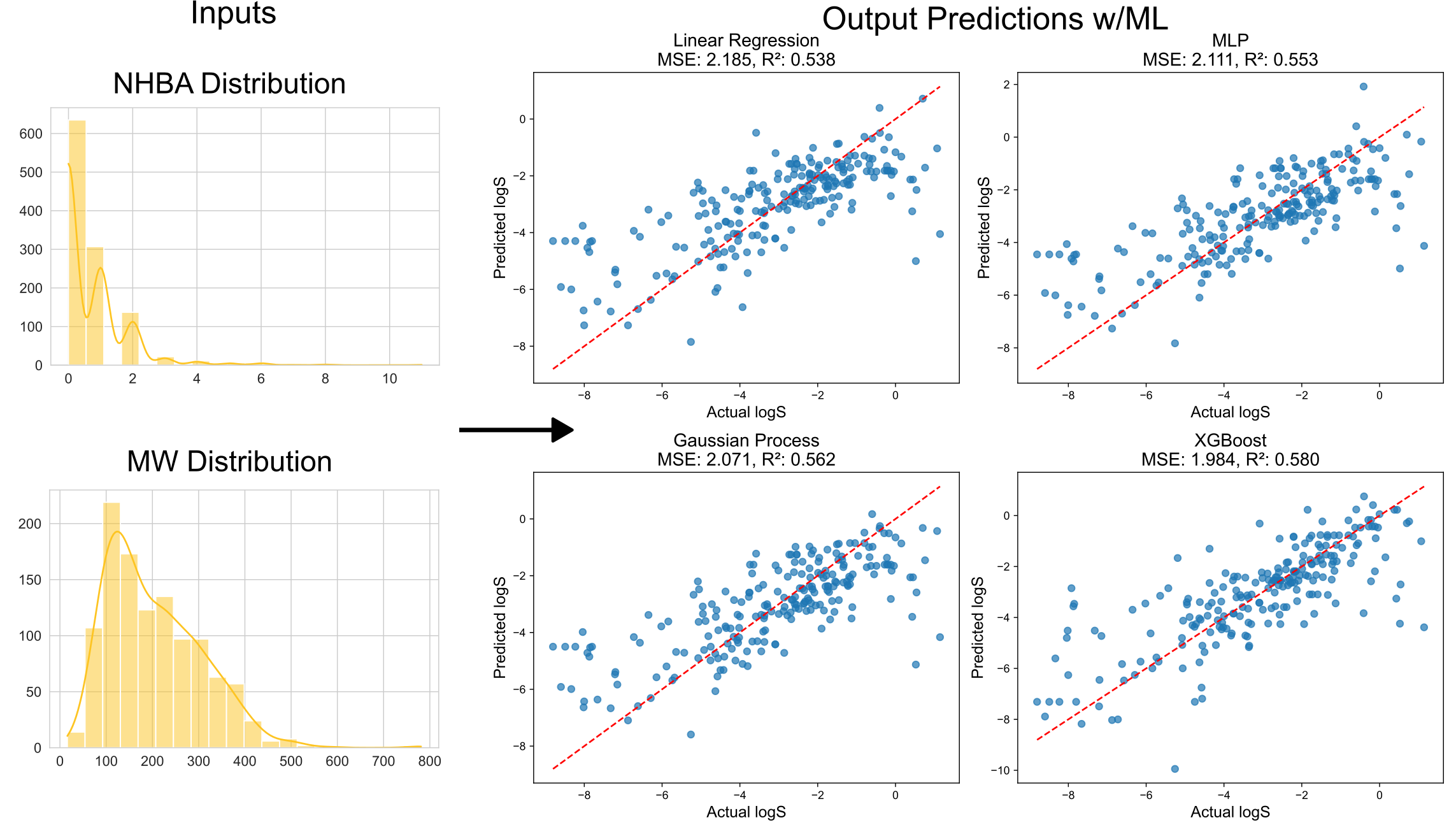}
  \mycaption{Traditional ML model performance comparison using $R^2$ as metric}{Showcase of two input features (NHBA = \# H-bond atom donors, MW = molecular weight) to predict logS using a linear regression model, Gaussian process, random forest, and an XGBoost model.}
  {}\label{Figure31}
\end{figure}

While a vast array of machine learning algorithms exists, we will focus on three representative examples: linear regression, Gaussian processes (GPs), and XGBoost (a popular implementation of gradient-boosted decision trees). Linear regression, due to its simplicity and direct connection to the architecture of neural networks, provides a foundational understanding of supervised learning. GPs offer strong predictive performance, particularly on smaller datasets, and, crucially, provide uncertainty estimates alongside their predictions, a valuable feature in scientific applications. XGBoost, a highly optimized implementation of gradient-boosted decision trees, is known for its robustness, scalability, and state-of-the-art performance on a wide range of tabular datasets.
To illustrate, consider predicting aqueous solubility (logS) using linear regression. Linear regression learns a linear transformation (a weighted sum of input features) that best fits the training data. In a highly simplified initial approach, we might use only molecular weight (MW) as a single input feature. The model's performance can be quantified using the R² metric, which measures the proportion of variance in the experimental logS values, explained by the model's predictions. A higher R² indicates a better fit, reflecting a stronger correlation between predicted and observed values. By systematically adding or modifying features and observing the impact on R², we can explore the importance of different molecular descriptors in predicting solubility.
Gaussian processes provide a probabilistic approach to regression, yielding both predictions and uncertainty estimates. A GP uses a kernel function to measure the similarity between molecules (represented by feature vectors). The closer two molecules are in feature space, the more correlated their predicted properties. This allows GPs to predict properties for new molecules based on training data, and to quantify prediction confidence. GPs excel in low-data regimes, common in chemistry, but their computational cost scales cubically with the number of data points, limiting their use on very large datasets. Figure \ref{Figure31} shows a comparison of these approaches.

\FloatBarrier

\subsection{Deep Learning and Neural Networks}

Having explored feature-based machine learning, we now turn to deep learning (DL), a subfield focused on neural networks. Neural networks are learnable, optimizable transformations of data, implemented as a series of parameterized operations across multiple layers. Data flows unidirectionally through the network, from input to output. Each layer's transformation is governed by learnable parameters (often called weights), that are iteratively adjusted during training to minimize a loss function. This optimization process allows the network to improve its performance on a given task. In essence, the network learns a hierarchy of representations, with each layer transforming the data into a form increasingly suited to the target task.
Multilayer Perceptrons (MLPs), also known as feedforward neural networks, are fundamental building blocks in deep learning. An MLP is structured as a series of interconnected layers, each composed of multiple processing units. While the initial concepts in neural networks drew inspiration from biological neurons, the field has significantly evolved, and the analogy is now primarily historical. Each connection between units in adjacent layers has an associated weight. Within each layer, the core operation is a linear transformation: the inputs to a unit are multiplied by their corresponding weights and summed. This weighted sum is then passed through a non-linear activation function. This combination of linear transformations and non-linear activations is crucial; it enables MLPs to approximate a wide range of complex, non-linear functions. The intermediate layers perform these successive transformations, while the final layer produces the network's prediction or output.
Returning to the logS prediction example, let's consider how an MLP processes the input data. Recall that in feature-based ML, we manually engineered features like molecular weight. An MLP, in contrast, learns to extract relevant features automatically. The input, which could be a feature vector describing multiple molecules (e.g., a 2x4 tensor representing two molecules with four features each), is first processed by the initial layer. This layer performs a linear transformation followed by a non-linear activation, resulting in a new representation, a hidden layer with, for example, 10 units (resulting in a 2x10 tensor in this case). Subsequent layers repeat this process, each applying its own linear transformation and non-linear activation to the output of the previous layer. The initial layers tend to learn low-level features, while deeper layers combine these into higher-level, abstract representations. This creates a hierarchy of learned features, optimized for predicting logS. The output of each hidden layer can be considered a point in a latent space – an abstract vector space where the learned features are encoded. Unlike feature-based methods where we hand-craft features, the MLP learns the optimal feature representation directly from the data during training.
Like MLPs, Graph Neural Networks (GNNs) learn hierarchical representations through successive learnable transformations. The crucial difference lies in the data they operate on: GNNs are designed for graph tensors, capturing the complex relationships inherent in graphs, while MLPs are limited to fixed-size vector inputs.

\subsection{Graph Neural Networks (GNNs)}

Graph Neural Networks (GNNs), as the name suggests, are neural networks designed to operate directly on graph-structured data. Put simply, if an MLP is a tensor-in, tensor-out transformation, a GNN is a graph-in, graph-out transformation. GNNs implement specialized operations to handle the unique properties of graphs – namely, their variable size, permutation invariance to node ordering, and relational structure. A GNN takes a graph tensor (as described in Chapter 1) as input and, through a series of transformations, produces a modified graph tensor as output. This output graph may have updated node, edge, and/or global features. Depending on the specific task (as discussed in Chapter 2), we might use only a subset of the output graph's components for decision-making. For example, when predicting a global property like LogP, we would primarily focus on the output global feature vector. The core mechanism enabling these transformations within a GNN is often described as message passing.
This mechanism enables GNNs to learn from graph-structured data. It's a process by which information is iteratively exchanged between different parts of a graph – nodes, edges, and the global context. This iterative exchange is what allows GNNs to handle graphs of varying sizes and structures, unlike MLPs, which require fixed-size inputs. Consider a molecular graph: an individual atom (represented as a node) has its own properties, but its behavior and role within the molecule are also profoundly influenced by its connections (edges) to other atoms and by the overall molecular context (global features). Message passing allows the GNN to integrate all this information. This process typically involves three key steps: message preparation, aggregation, and updating. Summarized as:
\begin{enumerate}
    \item \textbf{Preparation}: Gathering relevant feature information from the component being updated and its connected components. It is important to highlight that there is not always an aggregation step, and preparation can lead directly into updating.
    \item \textbf{Aggregation}: Combining information from multiple sources (if applicable) using a permutation-invariant function (e.g., sum, mean, max). 
    \item \textbf{Update}: Applying a learnable function (typically an MLP) to update the component's features based on the prepared information and, if applicable, the aggregated information.
\end{enumerate}

\begin{figure}[!htb] \centering
  \includegraphics[width=1.0\columnwidth]{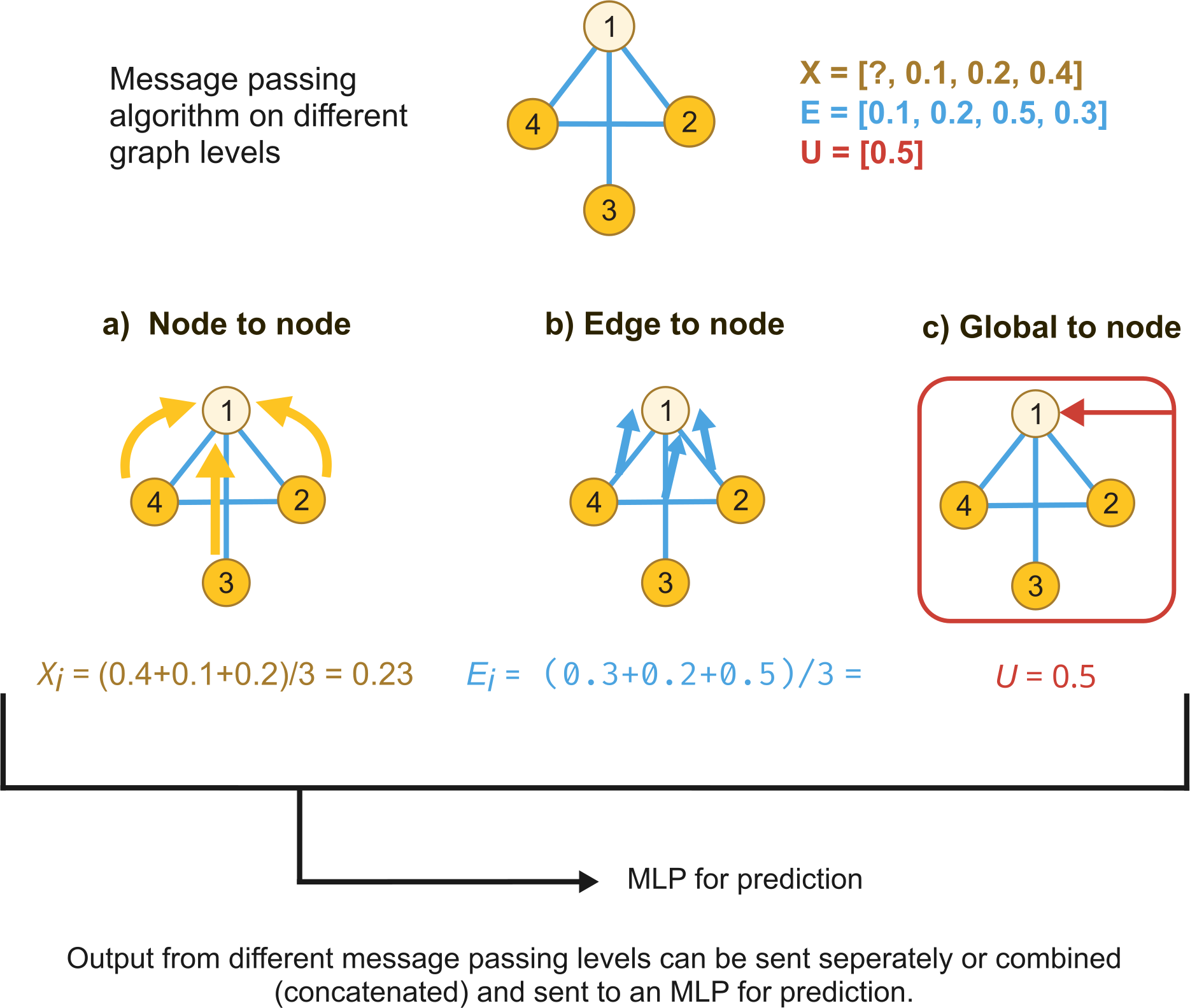}
  \mycaption{Message passing process for predicting node features}{Showcase of one method (mean) for passing different graph information to nodes for property prediction.
  }\label{Figure32}
\end{figure}
\FloatBarrier

\begin{figure}[!htb] \centering
  \includegraphics[width=1.0\columnwidth]{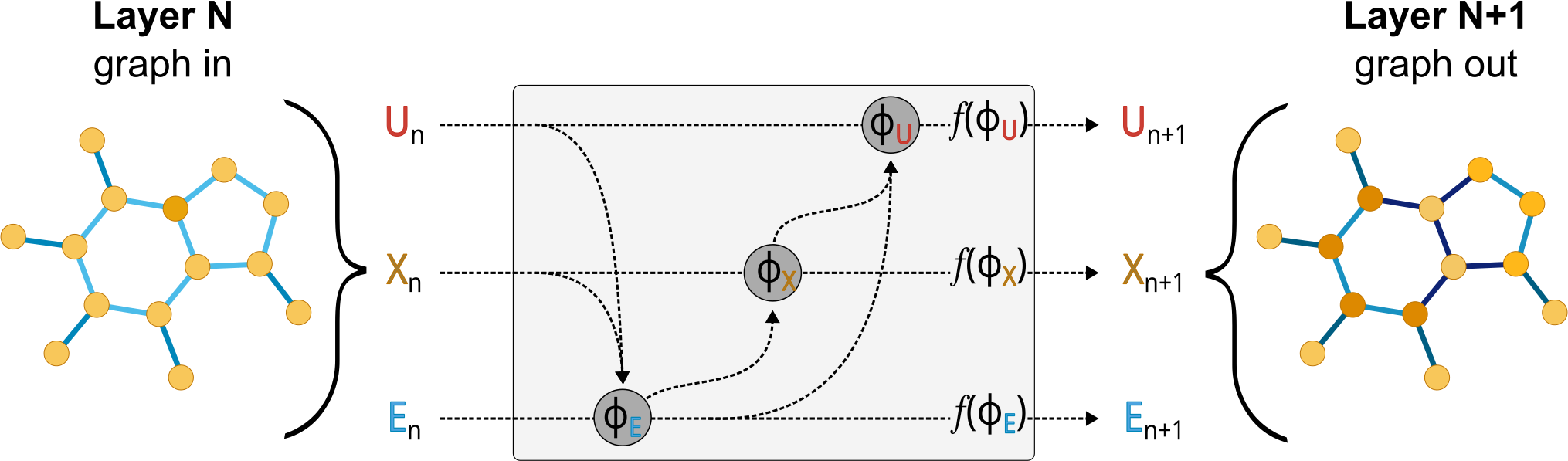}
  \mycaption{Graph Nets architecture ($\phi =$ aggregated vector and $f =$ update function)}{At each layer, edge features ($E_{n}$), node features ($X_{n}$), and global features ($U_{n}$) are aggregated ($\phi$) and updated ($f$) to produce the next-layer representations ($E_{n+1}, X_{n+1}, U_{n+1}$).
  }\label{Figure33}
\end{figure}

Numerous variations of message passing exist, leading to a diverse landscape of GNN architectures. Each variant differs in how it implements the message computation, aggregation, and update steps. However, many of these architectures can be understood as special cases of a more general framework called Graph Networks \cite{27}. The GraphNets framework explicitly considers all three components of a graph – nodes, edges, and globals – during message passing (Figure \ref{Figure33}). To illustrate the core concepts of message passing in a GNN, we will use a simplified, single-layer graph network applied to a methane molecule. 

\FloatBarrier

First let’s define the dimensions of each graph component tensor. We will have a node tensor of 5x3 (atomic number, partial charge, atom type) an edge tensor of 4x2 (bond type, dissociation energy) and a global tensor of 1x1 (molar mass). 
Within the GraphNets framework, message passing is performed sequentially in three distinct stages, called blocks: first, information is updated across edges; then, using the updated edge information, information is updated at the nodes; and finally, using all nodes and edges, the global features are updated.
Let's examine the edge block in detail, using our methane example. The first step is preparation. For each edge, we gather: the edge's own features (2 features), the features of the two nodes it connects (3 features each, totaling 6), and the global features (1 feature). These features are not yet aggregated; instead, we prepare them for input into a function. Next, for each edge, these collected features are concatenated into a single tensor. For a single edge, this results in a tensor of dimension 1x9. This concatenation is performed for every edge in the graph. Finally, an update function, typically a small MLP, is applied to each edge's concatenated feature tensor. This MLP transforms the 1x9 input tensor into a new updated edge feature tensor. If this MLP has 2 output units, the resulting updated edge feature tensor will have a dimension of 1x2. This process is applied independently to each edge in the graph.
Following the edge block, we move to the node block. First, we prepare the input for each node. For the carbon atom in our methane example, this involves gathering its initial 3 features, the four updated 1x2 edge feature tensors (from the edge block), and the 1x1 global feature tensor.
Next, we aggregate the updated edge features. Applying an aggregation function, a permutation-invariant function, such as sum, to the four 1x2 edge tensors, results in a single 1x2 aggregated edge feature tensor.
This 1x2 aggregated edge tensor is then concatenated with the node's original features (1x3) and the global features (1x1), resulting in a 1x6 tensor for the carbon atom.
Finally, an update function (a small MLP) transforms this 1x6 tensor into a new 1x3 updated node feature tensor. This process is repeated independently for every node in the graph.
Finally, we reach the global block, where the graph-level features are updated. The preparation step involves gathering: all the updated node feature tensors, all the updated edge feature tensors, and the current global feature tensor.
Next, we separately aggregate the updated node features and the updated edge features. A permutation-invariant function (e.g., sum, mean, max) is applied to the set of all updated node feature tensors, resulting in a single aggregated node feature tensor. Similarly, the same aggregation function is applied to the set of all updated edge feature tensors, producing a single aggregated edge feature tensor. In our methane example, if we use sum and assume there are five nodes with updated 1x3 feature tensors, the aggregated node tensor would be 1x3. If there are four updated 1x2 edge tensors, the aggregated edge tensor would be 1x2.
These aggregated node features (1x2), aggregated edge features (1x2), and the current global features (1x1) are then concatenated into a single tensor. In our methane example, this results in a 1x(3 + 2 + 1) = 1x6 tensor.
An update function (an MLP) then transforms this 1x6 tensor into an updated global feature tensor. If the MLP has 1 output unit, the final updated global feature tensor will be 1x1. This completes one full round of message passing in the GraphNets.
Each pass through the edge, node, and global blocks forms a GNN layer. Information propagates further with each layer. A single layer updates a node's features based on its immediate neighbors and the global features. An n-layer GNN allows information to travel up to n hops, meaning a 3-layer GNN captures a 3-hop radius around each node. The global features act as a central hub, facilitating long-range information transfer across the entire graph in a single layer, because they're updated using all nodes and edges, and then used in every node and edge update.

\begin{figure}[!htb] \centering
  \includegraphics[width=1.0\columnwidth]{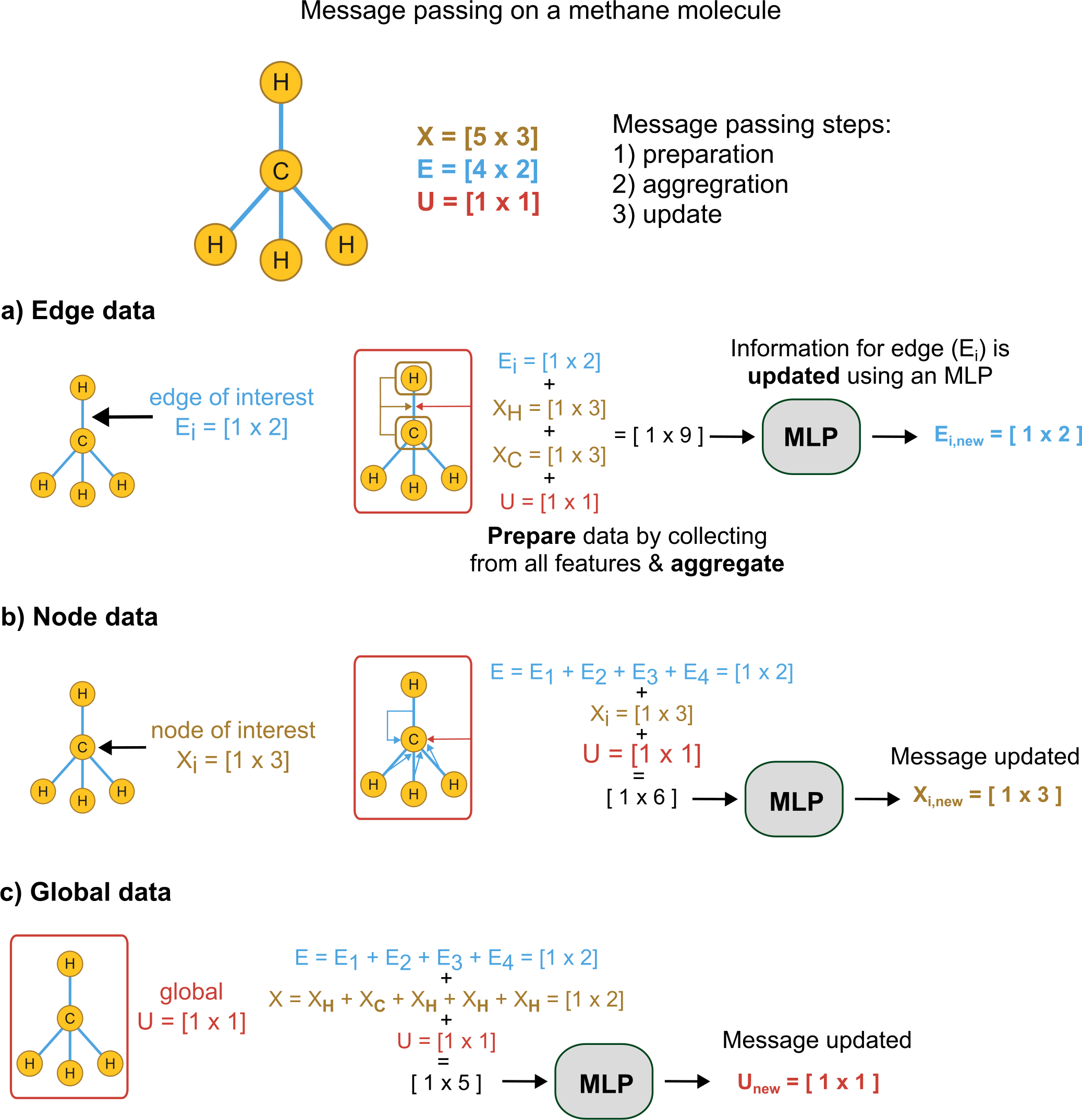}
  \mycaption{Message passing on methane on the edge, node, and global level}{Message passing on a methane molecule at the edge, node, and global levels. (a) Edge update: edge, connected node, and global features are concatenated and transformed by an MLP to yield updated edge features. (b) Node update: aggregated edge features, node features, and global features are combined and updated with an MLP. (c) Global update: aggregated edge and node features together with the global feature are concatenated and updated through an MLP to produce the new global feature.
  }\label{Figure34}
\end{figure}
\FloatBarrier

\subsection{The modelling space for GNNs}

The versatility of GNNs stems from the numerous design choices available when constructing a model. These choices primarily revolve around how the core operations of message passing – message preparation, aggregation, and updating – are implemented. Different choices lead to different GNN architectures, each with its own strengths and weaknesses.

\textbf{Key Design Decisions:}

\begin{enumerate}
    \item \textbf{Information Routing:} How is information routed between nodes, edges, and the global context? The Graph Networks framework, as previously discussed, provides a very general routing scheme, updating edges, then nodes, then globals. Other architectures might simplify this. For example, Graph Convolutional Networks (GCNs) focus solely on node-to-node message passing. They aggregate feature information from a node's immediate neighbors, typically normalizing by node degree to prevent nodes with high connectivity from dominating the aggregation. This can be viewed as a simplification of the Graph Networks node block, omitting edge and global features.
    \item \textbf{Aggregation Function:} How are messages from multiple sources (e.g., multiple neighboring nodes) combined? Common choices are permutation-invariant functions like sum, mean, max, and variance. More sophisticated aggregation strategies exist, such as Principal Neighbourhood Aggregation (PNA), which combines multiple aggregators to capture different aspects of the neighborhood distribution.
    \item \textbf{Message Computing:} How are messages generated? In the Graph Networks framework, the edge, node and global "messages" are just the concatenation of related features. Other methods compute messages. For example, Message Passing Neural Networks (MPNNs) provide a general framework where the message computation itself is a learnable function (often an MLP). This allows for greater flexibility in how information is exchanged between nodes and edges.
    \item \textbf{Update Function:} How are node, edge, and global features updated after receiving aggregated messages? While MLPs are commonly used, other differentiable functions could be employed.
    \item \textbf{Attention Mechanisms:} Should the influence of different neighbors be weighted equally, or should some neighbors be considered more important than others? Graph Attention Networks (GATs) introduce a learnable attention mechanism to weight the contributions of neighboring nodes during aggregation. Instead of fixed weights (as in GCNs), GATs learn attention coefficients that determine the relative importance of each neighbor's message.
    \item \textbf{Feature Conditioning:} How are different feature sources (e.g., node features and global features) combined? Simple concatenation is a common approach, but more sophisticated techniques exist. For example, Feature-wise Linear Modulation (FiLM) layers allow one feature source to modulate the influence of another, providing a more flexible way to integrate information.
    \item \textbf{Special Data Considerations:} How can we integrate additional data beyond the basic node, edge, and global features? For example, in molecular applications, we often have access to 3D atomic coordinates. Standard GNNs, as described so far, are not inherently designed to handle such geometric information. Specialized architectures, like E(3)-Equivariant Neural Networks (E(3)NNs), address this by incorporating equivariance to 3D rotations, translations, and reflections. This means that if the input coordinates are rotated, the output of the network transforms in a predictable and consistent manner, reflecting the inherent symmetries of 3D space. Similar considerations apply to other types of data, such as node or edge weights, continuous attributes, or even time-series data in dynamic graphs. Each data type may require specialized processing or architectural modifications to be effectively incorporated into the GNN.
\end{enumerate}

These design choices are not mutually exclusive; they can be combined in various ways. The optimal choices depend on the specific task and the characteristics of the graph data. By understanding these fundamental building blocks, we can appreciate the diversity of GNN architectures and tailor models to specific chemical problems.

\begin{figure}[!htb] \centering
  \includegraphics[width=1.0\columnwidth]{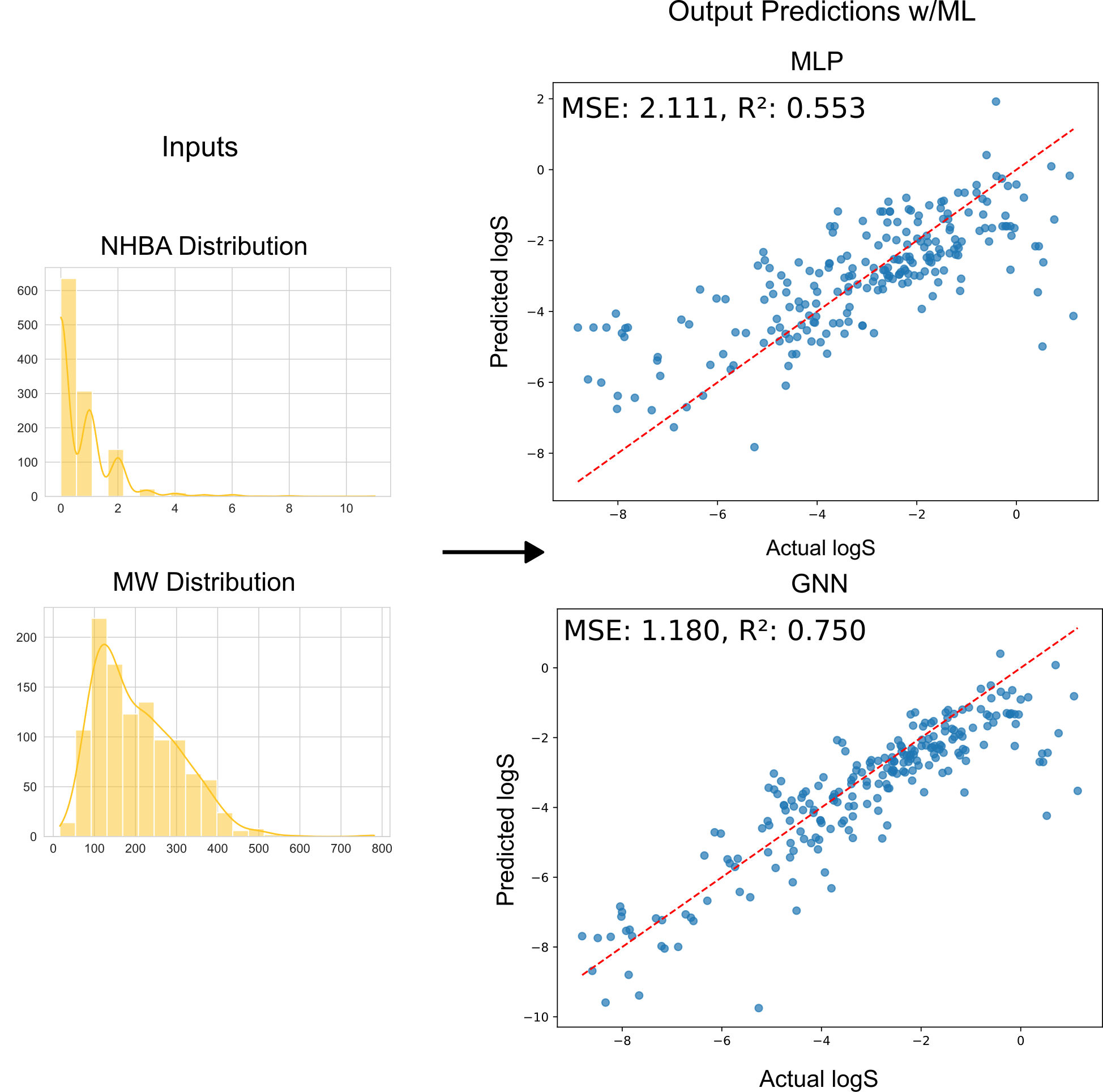}
  \mycaption{Comparison GNN vs MLP for predicting logS values for different molecules from input MW and NHBA}{Scatter plots show predicted versus actual logS values, with mean squared error (MSE) and coefficient of determination (R²) reported for each model. The GNN achieves lower error and higher predictive performance compared to the MLP.
  }\label{Figure35}
\end{figure}
\FloatBarrier

\section{5. Concluding remarks.}

From molecules, to proteins, to processes, graphs provide a natural and powerful way to represent chemical and biological systems. The application of graph-based methods, and particularly graph neural networks, is arguably more developed in chemistry than in many other scientific fields. This primer has equipped you with the foundational knowledge to understand and apply GNNs to a wide range of chemical problems. By mastering the concepts of graph representation (Chapter 1), task formulation (Chapter 2), diverse graph applications (Chapter 3) and the core principles of GNN operation (Chapter 4), you are now well-positioned to explore the rapidly growing literature, experiment with different GNN architectures, and contribute to the exciting advancements being made at the intersection of machine learning and chemistry.
While the potential is vast, challenges remain. The increasing scale and complexity of chemical systems can lead to computationally demanding graph representations \cite{28}. Modeling dynamic systems, where nodes and edges change over time, requires ongoing methodological development \cite{29}. And, in some areas, like drug discovery, data scarcity remains a significant hurdle \cite{30}.
Despite these challenges, the field is rapidly advancing. While chemistry has led the way in adopting graph-based approaches, the principles and techniques discussed here are broadly applicable. Significant opportunities exist to adapt and extend these methods to other scientific domains where relational data and complex systems are prevalent. The future of research is increasingly data-driven, and GNNs are a key technology for unlocking the potential of this data, both within chemistry and beyond.

\begin{suppinfo}
\section{Glossary of Terms}

\subsection{Chapter 1 - Graphs and Graph Data}
\textbf{Adjacency list:} A memory-efficient representation of graph connectivity that lists only the edges existing between nodes. \\ 
\textbf{Adjacency matrix:} A square matrix representing connections between nodes, where each entry indicates the presence or absence of an edge. \\
\textbf{Aggregation:} The process in graph-based models of summarizing information from neighboring nodes or edges to compute updated features for a given node, edge, or the entire graph.\\
\textbf{Algorithms:} Systematic computational procedures designed to operate on graph structures. These include methods for representing, analyzing, and manipulating graphs.\\
\textbf{Concatenation:} The operation of joining multiple vectors or tensors along a specified dimension to create a larger tensor. This is often used to combine different types of information or features in graph representations.\\
\textbf{Directed graph:} Type of graph where edges have a specific direction, meaning they point from one node to another rather than being bidirectional.\\
\textbf{Edges:} The fundamental components of a graph that represent the relationships or interactions between two nodes. They define how nodes are connected and can carry additional information depending on the application.\\
\textbf{Edge tensor:} Numerical representation of the features or properties associated with the edges in a graph. It encodes information about the relationships between nodes in a structured and computationally usable format.\\
\textbf{Global properties:} Attributes or features that are associated with the entire graph rather than individual nodes or edges.\\
\textbf{Global tensor:} A numerical representation that encapsulates the global properties of an entire graph. It serves as a single structure that aggregates information relevant to the graph as a whole, rather than specific nodes or edges.\\
\textbf{Graph:} Mathematical structure used to represent relationships or interactions between objects. It is composed of three main components: nodes (vertices), edges (links), and optionally, global properties.\\
\textbf{Heterogeneous graph:} Type of graph that contains multiple types of nodes and/or edges. This flexibility allows heterogeneous graphs to represent complex systems with diverse interactions and relationships.\\
\textbf{Hierarchical graph:} Type of graph where nodes themselves can represent subgraphs, allowing for a multi-scale or nested representation of complex systems. \\
\textbf{Homogeneous graph:} Type of graph where all nodes and edges are of the same type.\\ 
\textbf{Nodes:} Fundamental components of a graph that represent individual entities or objects within the system being modeled. Nodes are connected to one another via edges, and they can carry specific features or attributes that describe their properties.\\
\textbf{Node tensor:} Numerical representation of the features or attributes associated with the nodes in a graph. It organizes the properties of each node into a structured format that is suitable for computational analysis and machine learning tasks.\\
\textbf{One-hot encoding:} Method of representing categorical data as a numerical format that can be used in computational tasks.\\
\textbf{Rank and shape of a tensor:} Describe its structure and dimensions, which are crucial for representing data in a graph numerically. The rank of a tensor is the number of dimensions (axes) it has. The shape of a tensor is a tuple that specifies the size of the tensor along each dimension.\\
\textbf{Tensors:} Multidimensional arrays used to represent data numerically in graph-based computations. \\
\textbf{Graph tensor:} Refers to a collection of tensors that are organized to represent the components of a graph in a cohesive and computationally efficient manner. \\
\textbf{Undirected graph:} Type of graph where edges have no specific direction, meaning the relationship between nodes is bidirectional. Undirected graphs are commonly used to represent symmetric relationships or interactions in systems where the direction of the connection does not matter.\\
\textbf{Weighted graph:} Type of graph where edges are assigned weights, which represent additional quantitative information about the relationship or interaction between two nodes. \\

\subsection{Chapter 2 - What Kind of Problems Can We Solve With Graphs?}
\textbf{Edge-level tasks:} Computational problems that focus on predicting properties or features of the edges in a graph, or inferring the presence or absence of an edge. \\
\textbf{Feature engineering:} The process of enriching the input features of nodes, edges, or global graph components by adding additional, relevant information.\\
\textbf{Generative models:} Computational models designed to generate new graphs with specific properties or structures.\\
\textbf{Global-level tasks:} Computational tasks that focus on predicting properties or characteristics of the entire graph.\\
\textbf{Graph-level tasks:} Computational problems where the goal is to generate, modify, or analyze entire graphs.\\
\textbf{Interpretability tasks:} Involve analyzing graph components (nodes, edges, or global properties) to understand their contributions to a specific prediction or property.\\
\textbf{Inverse design:} The process of generating or optimizing graph structures (e.g., molecules, proteins, reaction pathways) to achieve specific desired properties or behaviors.\\
\textbf{Node-level tasks:} Focus on predicting or analyzing properties or attributes specific to individual nodes within a graph.\\

\subsection{Chapter 3 - Expanded Graph Applications}
\textbf{Equivariant 3D transformations:} The property of a model or graph representation to maintain consistency under transformations such as rotation, translation, or reflection, critical for preserving molecular 3D structural information.\\
\textbf{Molecular dynamics (MD) simulation:} Computational technique used to study and predict the behavior, properties, and interactions of molecules in various environments. It employs physics-based models and algorithms to simulate the dynamics, energetics, and structure of molecular systems\\
\textbf{Time-dependent graphs:} Graphs capturing temporal changes in systems by representing states at discrete time intervals or connecting nodes over time.\\

\subsection{Chapter 4 - Learning Algorithms on Graphs}
\textbf{Artificial intelligence:} Artificial intelligence (AI) is a branch of computer science focused on creating systems and algorithms that can perform tasks typically requiring human intelligence.\\
\textbf{Activation function:} A non-linear function applied to the output of a linear transformation in a neural network layer. This non-linearity allows neural networks to approximate complex functions.\\
\textbf{Aggregation function:} In the context of GNNs, a permutation-invariant function (e.g., sum, mean, max, variance) that combines information from multiple sources (e.g., neighboring nodes or edges) into a single representation.\\
\textbf{Decision tree:} A decision tree is a supervised learning algorithm used for classification and regression tasks. It models data using a tree-like structure, where each internal node represents a decision based on an attribute, branches denote outcomes, and leaf nodes provide final predictions.\\
\textbf{Descriptors:} Quantitative or qualitative representations of features used to characterize objects, data, or patterns in machine learning.\\
\textbf{Deep learning (DL):} A subfield of machine learning that uses neural networks with multiple layers (deep neural networks) to learn representations from data automatically.\\
\textbf{Ensemble learning:} Machine learning technique that combines multiple models to achieve better predictive performance than any individual model alone.\\
\textbf{Gaussian processes:} A probabilistic machine learning model that defines a distribution over functions, allowing it to make predictions with uncertainty estimation. It is widely used in regression and Bayesian optimization.\\
\textbf{Graph neural network (GNN):} A class of neural networks designed to operate directly on graph-structured data.\\
\textbf{Graph networks / GraphNets:} A general framework for GNNs that explicitly considers nodes, edges, and global features during message passing.\\
\textbf{Graph traversal (graph walking):} Algorithms that systematically explore a graph, collecting information about node connectivity and substructures.\\
\textbf{Graph convolutional network (GCN):} A type of GNN that performs node-to-node message passing, aggregating features from neighboring nodes with fixed weights (typically normalized by node degree).\\
\textbf{Graph attention networks:} Enhances the message-passing process by using an attention mechanism. Instead of treating all neighbors equally, GAT assigns different attention weights to different neighbors.\\
\textbf{Hidden layer:} Any layer in a neural network between the input and output layers.\\
\textbf{Input layer:} The first layer in a neural network, which receives the raw input data.\\
\textbf{Linear regression:} Supervised learning algorithm used for predicting a continuous dependent variable based on one or more independent variables. It assumes a linear relationship between the input features and the target variable.\\
\textbf{Message passing neural networks:} Graph neural networks (GNNs) that generalize deep learning methods to graph-structured data by iteratively exchanging and aggregating information between nodes and their neighbors.\\
\textbf{Message passing:} The core mechanism of GNNs, by which information is iteratively exchanged between different parts of a graph.\\
\textbf{Multi-Layer perceptron:} Class of feedforward neural networks composed of multiple layers of neurons. It consists of an input layer, one or more hidden layers, and an output layer.\\
\textbf{Machine learning (ML):} A field of computer science that enables computers to learn from data without being explicitly programmed.\\
\textbf{Neural network:} A computational model that performs learnable transformations of data. These transformations are implemented as a series of parameterized operations across multiple layers, allowing the network to learn complex patterns and relationships.\\
\textbf{Output layer:} The final layer in a neural network, which produces the model's prediction.\\
\textbf{Performance metrics:} Used to evaluate the performance of a machine learning model. The choice of metric depends on the type of problem.\\
\textbf{Weights:} Trainable parameters in a machine learning model that determine the strength of connections between neurons (or nodes). These weights are adjusted during training to minimize the difference between predictions and actual outputs.\\
\textbf{XGBoost:} Machine learning algorithm that constructs a series of decision trees in a sequential manner, where each tree improves upon the errors of the previous ones.\\

\end{suppinfo}
\clearpage

\bibliography{achemso-demo}
\providecommand{\latin}[1]{#1}
\makeatletter
\providecommand{\doi}
  {\begingroup\let\do\@makeother\dospecials
  \catcode`\{=1 \catcode`\}=2 \doi@aux}
\providecommand{\doi@aux}[1]{\endgroup\texttt{#1}}
\makeatother
\providecommand*\mcitethebibliography{\thebibliography}
\csname @ifundefined\endcsname{endmcitethebibliography}
  {\let\endmcitethebibliography\endthebibliography}{}

\end{document}